\documentclass[conference]{IEEEtran}
\IEEEoverridecommandlockouts
\usepackage{cite}
\usepackage{amsmath,amssymb,amsfonts}
\usepackage{graphicx}
\usepackage{textcomp}
\usepackage{xcolor}
\usepackage{caption}
\usepackage{subcaption}
\usepackage{multirow}
\usepackage{algorithm}
\usepackage{algorithmicx}
\usepackage{algpseudocode}
\algrenewcommand\algorithmicrequire{\textbf{Input:}}
\algrenewcommand\algorithmicensure{\textbf{Output:}}
\usepackage[a4paper, total={185mm,250mm}]{geometry}
\def\BibTeX{{\rm B\kern-.05em{\sc i\kern-.025em b}\kern-.08em
    T\kern-.1667em\lower.7ex\hbox{E}\kern-.125emX}}
\usepackage{comment}
\usepackage{authblk}
\usepackage{amssymb}

\usepackage[acronym]{glossaries} 
\usepackage[inline]{enumitem}
\newcommand{\todo}[1]{}
\renewcommand{\todo}[1]{{\color{red}TODO: {#1}}}
\newacronym{inn}{I-NN}{Inspect-NN}
\newacronym{mre}{MRE}{Mean Relative Error}
\newacronym{mae}{MAE}{Mean Absolute Error}
\newacronym{ai}{AI}{Artificial Intelligence}
\newacronym{aig}{AIG}{And-Inverter Graph}
\newacronym{adc}{ADC}{Approximate Don't Care}
\newacronym{aem}{AEM}{Average Error Magnitude}
\newacronym{alu}{ALU}{Arithmetic Logic Unit}
\newacronym{ama}{AMA}{Approximate Mirror Adder}
\newacronym{amosa}{AMOSA}{Archived Multi-Objective Simulated Annealing}
\newacronym{ann}{ANN}{Artificial Neural Network}
\newacronym{asic}{ASIC}{Application-Specific Integrated Circuit}
\newacronym{ast}{AST}{Abstract Syntax Tree}
\newacronym{axa}{AXA}{Approximate XOR-based Adder}
\newacronym{axc}{AxC}{Approximate Computing}
\newacronym{axct}{AxCT}{Approximate Computing Technique}
\newacronym{awce}{AWCE}{Absolute Worst-Case Error}
\newacronym{beb}{B \& B}{Branch \& Bound}
\newacronym{bdd}{BDD}{Binary Decision Diagram}
\newacronym{bmf}{BMF}{Boolean Matrix Factorization}
\newacronym{cl}{CL}{Convolutional Layer}
\newacronym{cnn}{CNN}{Convolutional Neural Network}
\newacronym{cmos}{CMOS}{Complementary Metal-Oxide Semiconductor}
\newacronym{cgp}{CPG}{Cartesian Genetic Programming}
\newacronym{dbx}{DBX}{Decision-BoX}
\newacronym{dct}{DCT}{Discrete Cosine Transform}
\newacronym{dfg}{DFG}{Data-Flow Graph}
\newacronym{dnn}{DNN}{Deep Neural Network}
\newacronym{dse}{DSE}{Design-Space Exploration}
\newacronym{dsp}{DSP}{Digital Signal Processing}
\newacronym{dssim}{DSSIM}{Structural DISSIMilarity}
\newacronym{dt}{DT}{Decision Tree}
\newacronym{dtmcs}{DT MCS}{Decision Tree based Mutiple Classifier System}
\newacronym{ea}{EA}{Evolutionary Algorithm}
\newacronym{e-idea}{E-$\mathbb{I}\text{\normalfont{DE}}\mathbb{A}$}{Evolutionary-IIDEAA Is a Design Exploration tool for Approximate Algorithm}
\newacronym{es}{ES}{Exact Synthesis}
\newacronym{eu}{EU}{Energy  Unit}
\newacronym{exdc}{ExDC}{External Don't Care}
\newacronym{fi}{FI}{Fault Injection}
\newacronym{fac}{FAC}{Full-Adder Cell}
\newacronym{fcl}{FCL}{Fully-Connected Layer}
\newacronym{fft}{FFT}{Fast Fourier Transform}
\newacronym{finfet}{FinFET}{Fin Field-Effect Transistor}
\newacronym{fir}{FIR}{Finite Impulse Response}
\newacronym{flap}{FLAP}{FLexible Arithmetic Precision}
\newacronym{fraig}{FRAIG}{Functionally-Reduced And-Inverter Graph}
\newacronym{fpga}{FPGA}{Field Programmable Gate Array}
\newacronym{fpu}{FPU}{Floating-Point Unit}
\newacronym{ga}{GA}{Genetic Algorithm}
\newacronym{gpgpu}{GP-GPU}{General Purpose - Graphic Processing Unit}
\newacronym{gpu}{GPU}{Graphic Processing Unit}
\newacronym{gmdh}{GMDH}{Group Method of Data Handling}
\newacronym{hdl}{HDL}{Hardware Description Language}
\newacronym{hls}{HLS}{High Level Synthesis}
\newacronym{iac}{IAC}{Inexact-Adder Cell}
\newacronym{ijcnn}{IJCNN}{International Joint Conference on Neural Networks}
\newacronym{inxa}{InXA}{IneXact Adder}
\newacronym{ilp}{ILP}{Integer Linear Programming}
\newacronym{llvm}{LLVM}{Low Level Virtual Machine}
\newacronym{lut}{LUT}{Look-Up Table}
\newacronym{mape}{MAPE}{Mean Absolute Percentage Error}
\newacronym{mlp}{MLP}{Multi-Layer Perceptron}
\newacronym{mnist}{MNIST}{Modified National Institute of Standards and Technology}
\newacronym{moea}{MOEA}{Multi-Objective Evolutionary Algorithm}
\newacronym{moco}{MOCO}{Multi-Objective Combinatorial Optimization}
\newacronym{mop}{MOP}{Multi-objective Optimization Problem}
\newacronym{mosfet}{MOSFET}{Metal-Oxide-Semiconductor Field-Effect Transistor}
\newacronym{mpl}{MPL}{Max-Pooling Layer}
\newacronym{mpcnn}{MP-CNN}{Max-Pooling Convolutional Neural-Network}
\newacronym{mpsoc}{MPSoC}{Multi-Processor System on Chip}
\newacronym{mse}{MSE}{Mean Squared Error}
\newacronym{mssim}{MSSIM}{Mean SSIM}
\newacronym{nab}{NAB}{Number of Approximate Bit}
\newacronym{nsgaii}{NSGA-II}{Non-dominated Sorting Genetic Algorithm-II}
\newacronym{noc}{NoC}{Network on Chip}
\newacronym{odc}{ODC}{Observability Don't Care}
\newacronym{pe}{PE}{Processing Element}
\newacronym{pi}{PI}{Primary Input}
\newacronym{pl}{PL}{Pooling Layer}
\newacronym{po}{PO}{Primary Output}
\newacronym{psnr}{PSNR}{Peak Signal-to-Noise Ratio}
\newacronym{qcc}{QCC}{Quality Constraint Circuit}
\newacronym{qec}{QEC}{Quality Evaluation Circuit}
\newacronym{relu}{ReLU}{Rectifier Linera Unit}
\newacronym{rem}{REM}{Relative Error Magnitude}
\newacronym{risc}{RISC}{Reduced Instruction-Set Computer}
\newacronym{rmse}{RMSE}{Root Mean Squared Error}
\newacronym{rnn}{RNN}{Recurrent Neural Network}
\newacronym{rtl}{RTL}{Register-Transfer Level}
\newacronym{sa}{SA}{Simulated Annealing}
\newacronym{sat}{SAT}{Boolean SATisfiability}
\newacronym{simd}{SIMD}{Single-Instruction-Multiple-Data}
\newacronym{simt}{SIMT}{Single-Instruction-Multiple-Thread}
\newacronym{smt}{SMT}{Satisfiability-Modulo Theory}
\newacronym{snr}{SNR}{Signal-to-Noise Ratio}
\newacronym{ssat}{\#SAT}{Sharp Boolean Satisfiability}
\newacronym{ssim}{SSIM}{Structural SIMilarity}
\newacronym{sqcc}{SQCC}{Sequential Quality Constraint Circuit}
\newacronym{vpa}{VPA}{Variable Precision Arithmetic}
\newacronym{wmed}{WMED}{Weighted Mean Error Distance}
\newacronym{ic}{IC}{Integrated Circuit}
\newacronym{ep}{EP}{Error Probability}
\newacronym{med}{MED}{Mean Error Distance}
\newacronym{ksa}{KSA}{Kogge-Stone adder}
\newacronym{cla}{CLA}{carry-lookahead adder}
\newacronym{rca}{RCA}{ripple-carry adder}
\newacronym{atm}{ATM}{array-tree multipler}
\newacronym{dtm}{DTM}{Dadda-tree multipler}
\newacronym{wtm}{WTM}{Wallace-tree multipler}
\newacronym{cska}{CSkA}{carry-skip adder}
\newacronym{hca}{HCA}{Han-Carlson adder}

\newcommand\blfootnote[1]{%
  \begingroup
  \renewcommand\thefootnote{}\footnote{#1}%
  \addtocounter{footnote}{-1}%
  \endgroup
}
\usepackage{fancyhdr}
\fancypagestyle{firstpage}
{
    \fancyhead[L]{\footnotesize © 2023 IEEE.  Personal use of this material is permitted.  Permission from IEEE must be obtained for all other uses, in any current or future media, including reprinting/republishing this material for advertising or promotional purposes, creating new collective works, for resale or redistribution to servers or lists, or reuse of any copyrighted component of this work in other works. This paper is accepted at the 41th IEEE VLSI Test Symposium (VTS) 2023.}
    \fancyhead[R]{}
}

\begin{document}
\IEEEoverridecommandlockouts
\IEEEpubid{\makebox[\columnwidth]{ 979-8-3503-4630-5/23/\$31.00 \copyright2023 IEEE \hfill} \hspace{\columnsep}\makebox[\columnwidth]{ }}

\title{Special Session: Approximation and Fault Resiliency of DNN Accelerators\\}
\author[1]{Mohammad Hasan Ahmadilivani}
\author[2]{Mario Barbareschi}
\author[2]{Salvatore Barone}
\author[3]{Alberto Bosio}
\author[4,1]{\\Masoud Daneshtalab}
\author[2]{Salvatore Della Torca}
\author[5]{Gabriele Gavarini}
\author[1]{Maksim Jenihhin}
\author[1]{\\Jaan Raik}
\author[5]{Annachiara Ruospo}
\author[5]{Ernesto Sanchez}
\author[1*]{Mahdi Taheri}

\affil[1]{Tallinn University of Technology, Tallinn, Estonia}
\affil[2]{University of Naples Federico II, Naples, Italy}
\affil[3]{Ecole Centrale de Lyon, Lyon, France}
\affil[4]{Mälardalen University, Västerås, Sweden}
\affil[5]{Politecnico di Torino, Torino, Italy}

\maketitle
\thispagestyle{firstpage}
\vspace{-0.2cm}\begin{abstract}

Deep Learning, and in particular, \gls{dnn} is nowadays widely used in many scenarios, including safety-critical applications such as autonomous driving. In this context, besides energy efficiency and performance, reliability plays a crucial role since a system failure can jeopardize human life. As with any other device, the reliability of hardware architectures running \glspl{dnn} has to be evaluated, usually through costly fault injection campaigns. 
This paper explores approximation and fault resiliency of \acrshort{dnn} accelerators.
We propose to use approximate (AxC) arithmetic circuits to agilely emulate errors in hardware without performing fault injection on the DNN. To allow fast evaluation of \acrshort{axc} \acrshort{dnn}, we developed an efficient \acrshort{gpu}-based simulation framework. 
Further, we propose a fine-grain analysis of fault resiliency by examining fault propagation  and masking in networks. 
\end{abstract}

\blfootnote{*The authors are sorted in alphabetic order.}

\begin{IEEEkeywords}
deep neural networks, approximate computing, fault emulation, reliability, resiliency assessment
\end{IEEEkeywords}

\section{Introduction}

Deep Neural Networks (DNNs) have evolved to be increasingly applied to assist different aspects of human life, e.g., healthcare, transportation, security, IoT and edge applications \cite{sze2017efficient}. In this context, energy efficiency and performance are the key constraints to be taken into account in designing \acrshort{dnn} accelerators. Approximate Computing (AxC) is an emerging paradigm applied for improving their efficiency that produces acceptable results despite inaccuracies in the computations \cite{armeniakos2022hardware,Bos2022}.

Employing \acrshort{dnn} accelerators in safety-critical applications has raised hardware reliability concerns. In compliance with ISO 26262 functional safety standard for road vehicles, the FIT (Failures In Time) rate of particular hardware components has to be 10 failures in 1 billion hours of operation at maximum to meet the target safety integrity level, which necessitates very circumspect design \cite{nardi2017functional, dft19}. 
The reliability of \acrshort{dnn} accelerators is boosted by their ability to function correctly even in the presence of environment-related faults (soft errors, electromagnetic effects, temperature variations) or faults in the underlying hardware (manufacturing defects, process variations, nanoelectronics aging effects) \cite{shafique2020robust}. \glspl{dnn} are known to be resilient to faults due to their numerous interconnected layers and the ability to mask faults \cite{bosio2019reliability}. However, several studies in recent years have shown that the accuracy of \glspl{dnn}  may still drop significantly in the presence of faults \cite{shafique2020robust,mittal2020survey,ibrahim2020soft,torres2017fault,su2023testability}. These observations demonstrate that the reliability of \acrshort{dnn} accelerators must be considered alongside efficiency. Some research works studied the reliability of approximated \glspl{dnn}  to show the trade-off between reliability and efficiency \cite{luza2020investigating, taheri2}.


The key challenge for DNN efficiency and reliability is the exploration of the huge design space. 
As mentioned, employing \acrshort{axc} units in \acrshort{dnn} accelerators is one of the eminent approaches to gaining efficiency. However, the design space for approximated \glspl{dnn}  is too large~\cite{Sek2023}, and implementing different \acrshort{axc} units to find an optimum efficiency is impracticable for FPGA accelerators. Notably, \glspl{gpu} that are widely applied for accelerating the \acrshort{dnn} training can be utilized to assist this process as well. To tackle the task of exploiting \acrshort{axc} in \glspl{dnn}, we present a \acrshort{gpu}-accelerated framework for \acrshort{dnn} approximation exploration.

Addressing accelerators' reliability issues starts with architecture-level fault-resiliency evaluation. \acrfull{fi} is a conventional method for this purpose that has been vastly applied for \glspl{dnn} as well \cite{ruospo2021pros,bosio2021emerging}. The main approaches for \acrshort{fi} experiments are fault simulation in software and fault emulation in hardware, both implying a huge fault space. Fast fault emulation in accelerators (especially in FPGAs, which are widely used for \glspl{dnn} \cite{talib2021systematic}) is still a challenge because of its iterative procedure, including numerous extra memory accesses as well as huge fault injection campaigns. To tackle this issue, we leverage \acrshort{axc} units in \glspl{dnn} as a non-conventional use of both FI and \acrshort{axc}, to emulate errors in the accelerator hardware. In this method, \acrshort{axc} units and their variants are a substitution for \acrshort{fi} targeting the fault resilience analysis of \acrshort{dnn} architectures.

Moreover, reducing fault space can also be done at the software level. We have carried out an empirical study on the inherent resilience to faults and errors of \glspl{dnn}, with the aim of investigating how they can mask a large portion of faults. In line with this, we propose the adoption of three different metrics to compute in advance (right after the injection of the fault) the effect the fault will have on the output vector score. In this way, it might be possible to both reduce the fault space and lower the \acrshort{fi} time.

The paper is organized as follows: Section \ref{sec:AxC-framework} introduces the GPU-accelerated framework for \glspl{dnn}  approximation exploration, Section \ref{sec:AxC-FI} presents a method for harnessing approximation for agile analysis of fault resiliency in \acrshort{dnn} accelerators, Section \ref{sec:fault-mask} provides a fine-grain \glspl{dnn} fault resiliency study by examining fault propagation and masking in networks, and Section \ref{sec:conclusion} concludes the paper.

\section{GPU Accelerated Framework for CNN Approximation}\label{sec:AxC-framework}

\subsection{Motivations and Related Works}
As stated in the introduction, the Approximate Computing paradigm is widely used to improve the energy efficiency of hardware accelerators for \glspl{dnn}. In particular, one promising solution is to use approximate arithmetic circuits~\cite{barbareschi_advancing_2021, barbareschi_genetic-algorithm-based_2022, barbareschi_catalog-based_2022}. 
However, quantifying the error introduced by these circuits requires expensive hardware prototyping, and, as a result, a software emulator of the \gls{dnn} accelerator is often executed on a CPU or \gls{gpgpu} instead. 
Nevertheless, this emulation is typically much slower than a software \gls{dnn} implementation running on a CPU or \gls{gpgpu} that uses the standard floating-point arithmetic instructions and common \gls{dnn} libraries because CPUs and \glspl{gpgpu} lack hardware support for approximate arithmetic operations; therefore, the latter operations  must be emulated, that is costly.

To address this issue, we propose \gls{inn}, that provides efficient emulation for approximate circuits to be deployed in \glspl{dnn} accelerator: approximate circuits are implemented as look-up tables and accessed through the memory mechanism of CUDA-capable \glspl{gpgpu}, reducing the inference time of the emulated \gls{dnn} accelerator by approximately 200 times compared to an optimized CPU version on complex \glspl{dnn}.

In the following, we present the \gls{inn} framework in Section~\ref{sec:inspectnn}, while Section~\ref{sec:inspectnn-case-studies} discusses case studies concerning the use of the mentioned framework to assess the accuracy loss due to approximate multipliers in \glspl{ann}.

\subsection{Proposed method}\label{sec:inspectnn}

The main purpose of the \gls{inn} framework is to investigate the impact of erroneous components on \gls{ai} applications.
In particular, it allows investigating how the accuracy of \glspl{dnn}-based applications is affected by imprecise components, i.e., those that do not meet their nominal behavioral specifications either because of faults, or because they have been specifically designed to differ in a controlled way from that behavior, while pursuing performance advantages. Examples are arithmetic components designed while exploiting the \gls{axc} design paradigm~\cite{bosio_approximate_2022}.
The behavior of imprecise components are modeled at the behavioral level by exploiting lookup tables, in which input operands select the corresponding output of the component. 
\gls{inn} exploits parallelism allowed by \glspl{gpgpu}: the inference phase is split in blocks, each assigned to a thread block on the \gls{gpgpu} and executed independently and parallelly from the others. 
\gls{inn} does the latter computation through a kernel, i.e., a CUDA function called by the CPU and executed on the \gls{gpgpu}: operations within each layer are parallelized so that each thread block execute a part of the overall operation; then, if needed, the output is normalized to be represented using \textit{n} bits, with \textit{n} being configurable.
Data exchange between the CPU and the \gls{gpgpu} are minimized: data is copied from the \gls{gpgpu} memory to the CPU ones when strictly required; hence, if two consecutive layers are working on the \gls{gpgpu}, the first one feeds the \gls{gpgpu} memory address of the computed data to the next layer, rather than coping them back and forth from/to the CPU.

\subsection{Experimental Results}\label{sec:inspectnn-case-studies}

\begin{table}[]
	\caption{Error characterization and hardware requirements for approximate circuits taken from the EvoApproxLib-Lite library, as reported in~\cite{mrazek_scalable_2018}}
	\label{tab:evoapprox-multipliers}
	\centering
	\resizebox{\columnwidth}{!}{%
		\begin{tabular}{|c|ccc|cc|}
			\hline
			\textbf{Circuit name} & \textbf{\begin{tabular}[c]{@{}c@{}}MAE\\ (\%)\end{tabular}} & \textbf{\begin{tabular}[c]{@{}c@{}}AWCE\\ (\%)\end{tabular}}& \textbf{\begin{tabular}[c]{@{}c@{}}MRE\\ (\%)\end{tabular}}& \textbf{\begin{tabular}[c]{@{}c@{}}Power\\ (nW)\end{tabular}} & \textbf{\begin{tabular}[c]{@{}c@{}}MAE\\ ($\mu m^2$)\end{tabular}}  \\ \hline
			mul8s\_1KV6 & 0.00  & 0.00  & 0.00  & 0.425 & 729.8 \\
			mul8s\_1KV8 & 0.0018  & 0.0076  & 0.28  & 0.422 & 711.0 \\
			mul8s\_1KV9 & 0.0064  & 0.026  & 0.90  & 0.410 & 685.2 \\
			mul8s\_1KVA & 0.019 & 0.075 & 2.53 & 0.391 & 641.1 \\
			mul8s\_1KVM & 0.049 & 0.20 & 2.40 & 0.369 & 652.8 \\
			mul8s\_1KVP & 0.051  & 0.21  & 2.73  & 0.363 & 635.0 \\
			mul8s\_1KVQ & 0.056 & 0.25 & 3.64 & 0.351 & 599.8 \\
			mul8s\_1KX5 & 0.15 & 0.69 & 8.93 & 0.289 & 543.0 \\
			mul8s\_1KXF & 0.34 & 1.37 & 15.72 & 0.237 & 482.4 \\
			mul8s\_1L2J & 0.081 & 0.39 & 4.41  & 0.301 & 558.9 \\
			mul8s\_1L2L & 0.23 & 1.16 & 12.26 & 0.200 & 411.6  \\
			mul8s\_1L2N & 0.52  & 2.66  & 27.44  & 0.126 & 284.9 \\
			mul8s\_1L12 & 3.08 & 12.30 & 135.77 & 0.052 & 172.2 \\ \hline
		\end{tabular}%
	}
 \vspace{10pt}
\end{table}

Case studies discussed in this Section concern the evaluation of the accuracy loss due to the use of multipliers taken from  the EvoApproxLib-Lite~\cite{mrazek_scalable_2018} library of approximate circuits while targeting several pre-trained \glspl{dnn}.
In particular, through \gls{inn}
\begin{enumerate*}[label=(\roman*)]
    \item we import the \gls{dnn} to be analyzed directly from the most common machine learning frameworks, such as TensorFlow, TensorFlow LITE, and 
    \item we define which specific approximate components have to be used, and
   \item we specify whether the analysis has to be performed at either coarse or fine grain.
\end{enumerate*}
In coarse grain analysis, a single approximate component is deployed in the whole network. 
Conversely, in fine-grain analysis, each layer of the target \gls{dnn} can use a different imprecise component.

We deploy multipliers from~\cite{mrazek_scalable_2018} --  whose error characterization and hardware overhead are reported in Table~\ref{tab:evoapprox-multipliers}, for the reader convenience -- to LeNet5 \gls{cnn}~\cite{lecun_gradient-based_1998}, to MinNet, and to ResNet-8~\cite{he_deep_2016}, that, although trained using floating-point arithmetic, are all quantized to use 8-bit integer.
The first \gls{cnn}, i.e., LeNet5,  has been trained to classify images from the \gls{mnist} benchmark~\cite{lecun_mnist_1998}, on which it exhibits 99.07\% accuracy. 
The MinNet \gls{cnn} is a custom-made \gls{cnn} inspired by the LeNet5 architecture: as for the latter, it consists of two \glspl{cl}, a \glspl{fcl} and one \glspl{pl} between each \gls{cl}, and it consists of approximately 160 thousand parameters. Despite its small size w.r.t. state-of-the-art networks, it  exhibits 80.07\% accuracy on the CIFAR-10 dataset~\cite{krizhevsky_cifar-10_2010}.
Last, the ResNet-8 \gls{cnn}, instead, has been trained while targeting images taken from the CIFAR-10 dataset~\cite{krizhevsky_cifar-10_2010}, which consists of 60 thousand RGB images, each belonging to one among ten classes. The network, that consists of more than 300 thousand learned parameters, and it exhibits 84.31\% accuracy on the mentioned dataset.
During the inference phase, these three architectures require performing 400 thousand, 4 million and 40 million multiplications each, respectively; hence, they represent a good test case for the evaluation of execution time.

\begin{table}[]
\caption{Accuracy loss and computational time for approximate circuits taken from the EvoApproxLib-Lite library~\cite{mrazek_scalable_2018}.}
\label{tab:inspectnn-evoapprox-acc-loss}
\centering
\resizebox{\columnwidth}{!}{%
\begin{tabular}{|c|ccc|cc|cc|}
\hline
\textbf{}             & \multicolumn{3}{c|}{\textbf{LeNet5}}                                                                                                                                                             & \multicolumn{2}{c|}{\textbf{MinNet}}                                                                                              & \multicolumn{2}{c|}{\textbf{ResNet8}}                                                                                             \\
\textbf{Circuit Name} & \textbf{\begin{tabular}[c]{@{}c@{}}Acc.\\ Loss\\  (\%)\end{tabular}} & \textbf{\begin{tabular}[c]{@{}c@{}}GPU\\ Time\end{tabular}} & \textbf{\begin{tabular}[c]{@{}c@{}}CPU\\ Time\end{tabular}} & \textbf{\begin{tabular}[c]{@{}c@{}}Acc.\\ Loss\\ (\%)\end{tabular}} & \textbf{\begin{tabular}[c]{@{}c@{}}GPU\\ Time\end{tabular}} & \textbf{\begin{tabular}[c]{@{}c@{}}Acc.\\ Loss\\ (\%)\end{tabular}} & \textbf{\begin{tabular}[c]{@{}c@{}}GPU\\ Time\end{tabular}} \\ \hline
mul8s\_1KV6 &  0 & 13.23s & $\approx$10h &  0 & 13.0s & 0 & 31.07s\\
mul8s\_1KV8 & 0.07 & 13.19s & $\approx$10h & -0.3 & 13.6s &  -0.19 & 31.1s\\
mul8s\_1KV9 & 0.15 & 13.27s & $\approx$10h &  0.3 & 13.6s &  -0.42 & 31.3s\\
mul8s\_1KVA & 0.51 & 13.22s & $\approx$10h &  2.5 & 13.5s & -0.08 & 31.3s\\
mul8s\_1KVM & 0.16 & 13.23s  & $\approx$10h &  -0.4 & 13.5s & 0.12 & 31.5s\\
mul8s\_1KVP & 0.27 & 13.17s & $\approx$10h &  -0.8 & 13.7s & -0.18 & 31.4s\\
mul8s\_1KVQ & 0.61 & 13.18s & $\approx$10h &  0.5 & 13.5s & 0.09 & 31.4s\\
mul8s\_1KX5 & 1.77 & 13.18s & $\approx$10h &  5.5 & 13.5s &  5.48 & 31.3s\\
mul8s\_1KXF & 1.57 & 13.18s & $\approx$10h &  -1.2 & 13.6s & 8.45 & 31.2s\\
mul8s\_1L2J & 0.79 & 13.2s & $\approx$10h &  46.6 & 13.6s &  74.61 & 31.5s\\
mul8s\_1L2L & 3.81 & 13.14s & $\approx$10h &  61.5 & 14.2s &  73.73 & 31.8s\\
mul8s\_1L2N & 15.92 & 13.11s & $\approx$10h &  65.9 & 14.0 s &  74.52 & 32.06s\\
mul8s\_1L12 & 75.66 & 13.15s & $\approx$10h &  66.4 & 14.4s & 74.49 & 33.6s \\ \hline
\end{tabular}%
}
\vspace{10pt}
\end{table}

To estimate the error introduced by the approximation, we execute the approximate \gls{cnn} to obtain its classification accuracy on the whole test data set, reporting the accuracy-loss and computational time required for the inference phase in Table~\ref{tab:inspectnn-evoapprox-acc-loss}.
The latter table also reports the error and hardware parameters for each of the considered approximate multipliers. 
We performed the inference phase on an NVIDIA RTX A5000~\gls{gpgpu}, that is built on the NVIDIA Ampere architecture and combines 256 Tensor Cores and 8192 CUDA cores with 24 GB of graphics memory.
Furthermore, for comparison purpose, the computational time of the inference phase while resorting to a CPU-only implementation is reported in Table~\ref{tab:inspectnn-evoapprox-acc-loss}. In this case, we leverage two 3.20 GHz Intel Xeon Silver 4210 CPUs, providing 20 cores / 40 threads computing power. We reported CPU time only for the LeNet5 case. For the  MinNet and ResNet8, the CPU execution time was higher than 10 hours and we were not able to complete the experiments.

As it is easy to foresee, the speed-up provided by the \gls{gpgpu} is crucial:  we can state that by exploiting the \gls{gpgpu} through our look-up table implementation of approximate multiplier allows for tremendous performance improvements, even though we compared the execution time. Furthermore, it can be noticed that the execution time increases as the number of multiplications performed during the inference phase increases, and it is independent of the particular approximate multiplier being deployed, as it can be observed in Table~\ref{tab:inspectnn-evoapprox-acc-loss}.



\section{Harnessing Approximation for Fault Injection in DNN Accelerators} \label{sec:AxC-FI}

\subsection{Motivations and Related Works}
\begin{figure*}
    \centering
    \includegraphics[width = 0.7\textwidth]{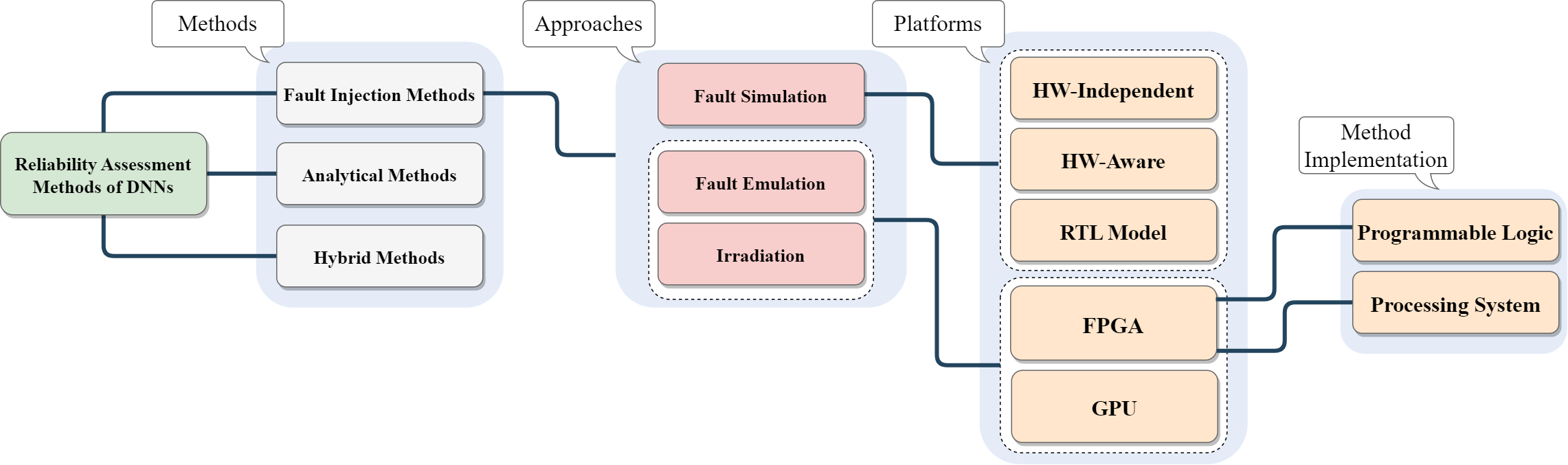}
    \caption{A taxonomy of DNN reliability assessment methods}
    \label{assess}
\end{figure*}
 
A major consequence of single or multiple accumulated soft-error-caused bitflips affecting the weights of a given layer is their propagation as errors at the layer outputs (also known as layer Output Feature Map) and further throughout the subsequent layers, leading to incorrect DNN predictions. \textit{Fault resilience} is the ability to tolerate the impact of faults on the output accuracy, and, in practice, it is one of the contributors to the final DNN accelerators’ reliability. A relevant mitigation strategy at the architecture level can be a hardening of the DNN, e.g., by layer redesign or selective hardening of neurons, such as hardened Processing Elements (PEs) or Triple Modular Redundancy (TMR) variants \cite{mittal2020survey}. These imply the assessment of layers' fault resiliency or identification of critical neurons in a neural network that are the most vulnerable to faults \cite{schorn2018accurate,ruospo2021reliability,ahmadi2023Deepvigor}. 
Fig. \ref{assess} presents a taxonomy for DNN reliability assessment methods. Along with analytical and hybrid methods \cite{ahmadi2023Deepvigor}, Fault Injection (FI) is a commonly used method for evaluating the fault resilience of DNNs \cite{su2023testability,ruospo2023survey,taheri1}. The industry often employs fault injection by emulation in hardware, particularly in FPGAs, as it allows for evaluating real-scale DNN accelerator designs in significantly shorter run times than software-based simulations \cite{ibrahim2020soft}. 

Fiji-FIN \cite{fiji} is a representative framework implemented on the embedded Processing System for evaluating the resiliency of DNNs by emulating FI on FPGA. It measures accuracy degradation as a metric to study the impact of soft errors on network parameters. Designing fault injection campaigns for such frameworks requires significant effort, as each injection halts inference execution to manipulate DNN parameters. This interrupts classification time for a batch of inputs.

The state-of-the-art approaches for FI by emulation in FPGA using the embedded Processing System often require iterative procedures for each injected fault. In particular, such an iterative approach breaks the pipeline execution of the accelerator, requires a complex FI controller, and needs an extra FI control interconnection to handle the injection \cite{hsueh1997fault, khoshavi2020shieldenn, fiji}. These procedures also involve multiple additional memory accesses, resulting in time-consuming processes and complex implementation. 

Unlike the works mentioned above, our proposed method can be classified as fault injection by emulation in Programmable Logic. It leverages the functional approximation as a substitute for the errors generated by FI to improve processing and design time as well as the control complexity in the DNN fault resiliency analysis process. This approach allows the inference pipeline to be executed on a batch of inputs without interruption. This agile method enables a fast and efficient exploration of different options for network architecture, training, dataset selection, and more, to study the fault resilience of DNNs. Specifically, the introduced errors mimic single or multiple accumulated faults in weights. The method allows for efficient analysis of how subsequent layers in the network tolerate errors in the Output Feature Map of an assumed compromised layer
are affected by faults in the weights of a compromised layer.

To the best of our knowledge, this is the first time that AxC units are utilized to enhance the efficiency and reduce the complexity of resilience analysis for DNNs.

\subsection{Proposed method}
\label{pr}
\begin{figure}
    \centering
    \includegraphics[width = 0.4\textwidth]{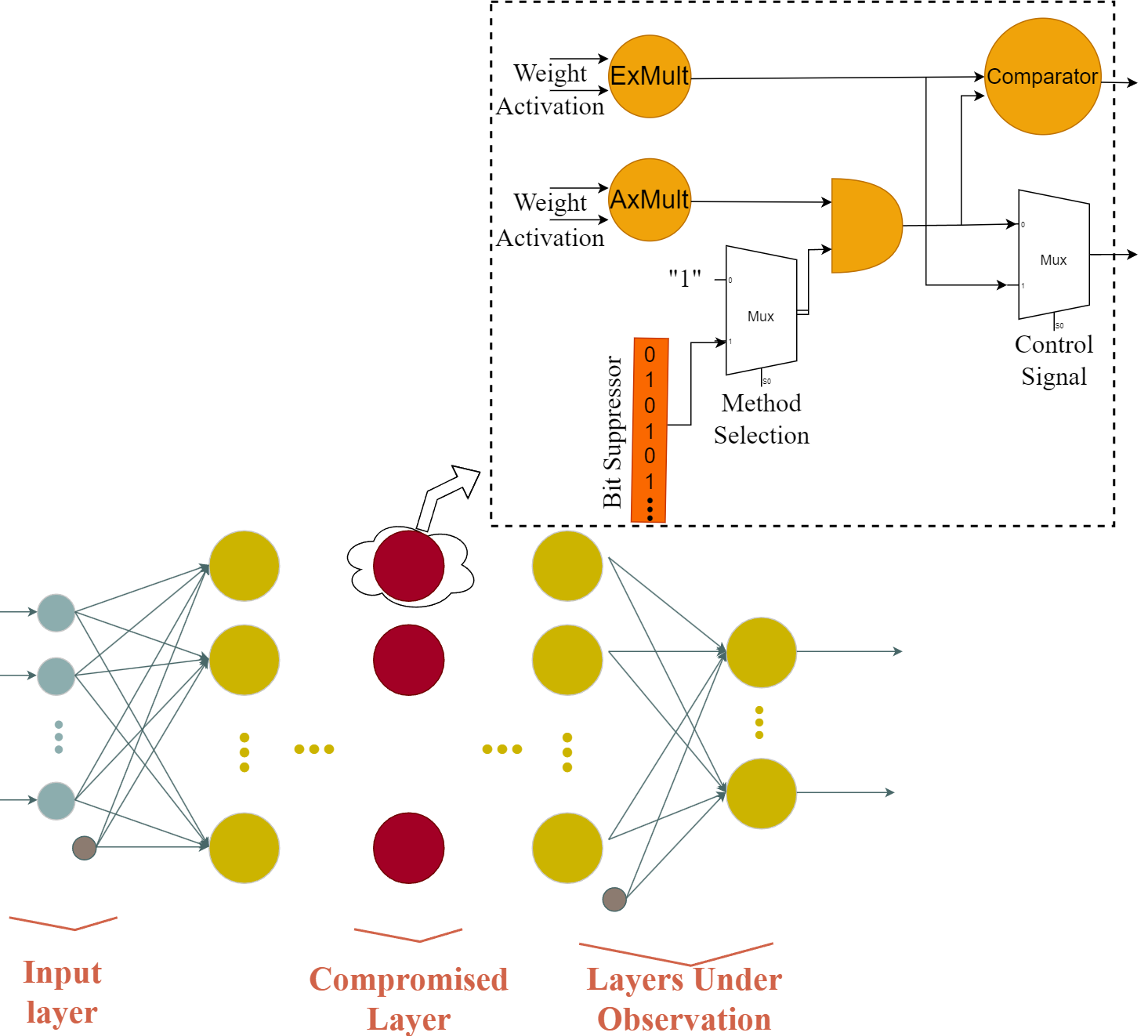}
    \caption{Proposed method evaluation}
    \label{experiment}
\end{figure}

AxC is commonly used to approximate hardware components to improve compute efficiency while maintaining functional accuracy. However, in practice, the errors induced by approximation can be used to mimic the errors caused by faults in logic circuits. These errors affect the outputs of the corresponding units and propagate to subsequent layers, impacting their activations (Fig. \ref{experiment}).
The proposed approach for evaluating DNN's fault resiliency using approximate computing (AxC) units is presented in Fig. \ref{experiment}. To implement our proposed method, an AxMult, or an AxMult + a bit suppression unit (AxMult+) is implemented along with the exact implementation of the multipliers (ExMult) in the network, depending on whether the network is being run in functional or fault resilience assessment mode. The golden inference for the validation dataset is run only once, and the layer outputs are stored and compared with a Comparator unit. The Bit Suppressor unit is meant to increase the probability of more significant bits of the neuron being impacted by faults. The less significant bits of the layer Output Feature Map are already affected by the AxMult with proper randomness depending on the data distribution in the network and layers.

\begin{figure}[h]
    \centering
    \includegraphics[width=0.5\textwidth]{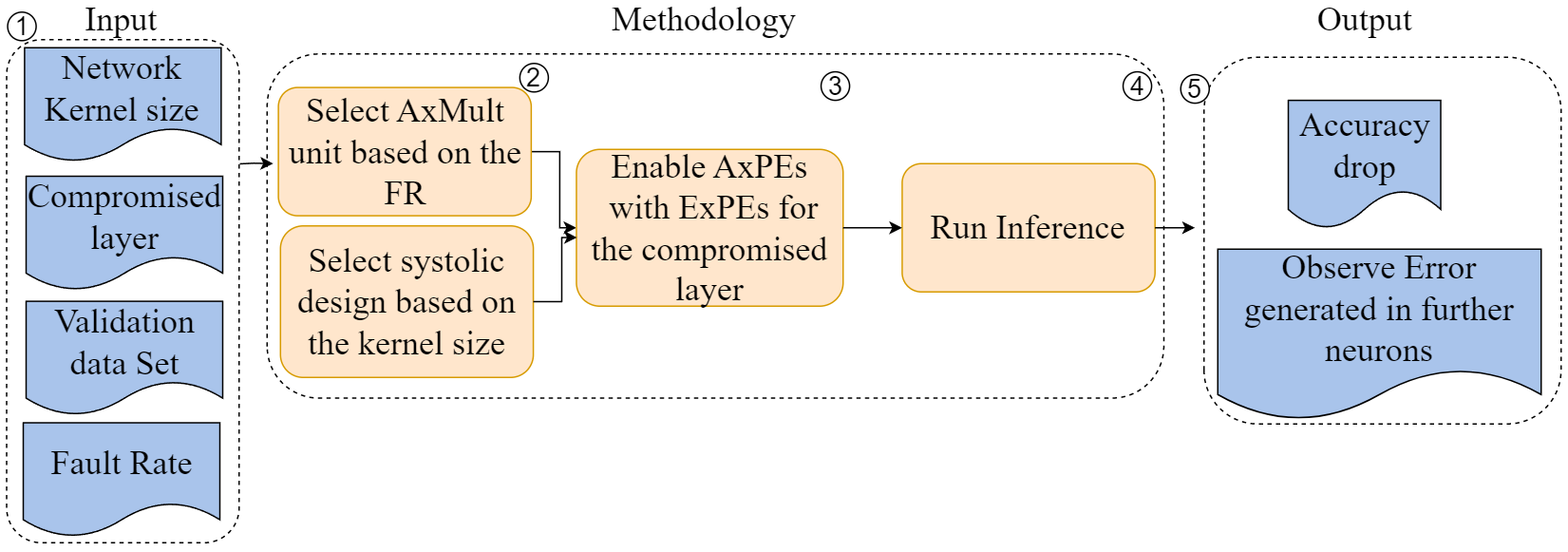}
    \caption{Methodology flow}
    \label{method}
\end{figure}

The overall flow of the proposed method is illustrated in Fig. \ref{method}. In Step 1, the user initializes the method by selecting the compromised layer in the DNN structure, the validation dataset (i.e., DNN inputs), and the application-specific target fault rate assumed for the analysis. In Step 2, suitable AxC units are selected for Approximate Processing Elements (AxPEs), such as the AxC multipliers from a relevant library, e.g., the EvoApproxLib \cite{evoapprox16}, or their variants with bit suppression. In Step 3, the selected AxMults started executing the compromised layer by enabling corresponding AxPEs along with the Exact Processing Elements (ExPE) in the DNN architecture. The DNN inference is run while keeping the network pipeline intact, and the resulting DNN output accuracy drop is recorded as the primary metric for analyzing DNN fault resilience. A more significant drop in accuracy with induced errors implies a less fault-resilient DNN implementation. At the same time, the outputs of the AxMults are compared with the ExMults outputs to calculate the actual error at each neuron. The rest of the inference is executed by ExMults for both erroneous and exact outputs, and the comparison is performed for all the subsequent neurons of the network.

The characteristics of the approximation-induced errors can be evaluated using different metrics such as normalized error, number of flipped bits, and impact on the neural network classification accuracy drop. In this study, we rely on a simple set of metrics that includes:

\begin{itemize}

\item Normalized error: the average error on the output of each layer is calculated by subtracting the neurons' outputs of that layer from the golden output and dividing all the error values by the maximum value.
\item Network accuracy: calculated by executing the network under different circumstances (faulty, AxMult, AxMult + bit suppressor and bit suppressor) over the test set.
\item Bitflips in subsequent layers: calculated by comparing all bits in the next layers' outputs with the golden model and counting the bits that do not match as flipped bits.
    
\end{itemize}
\subsubsection{Accelerator Model} \label{method:accelerator}

\begin{figure}[h]
    \centering
    \includegraphics[width=0.45\textwidth]{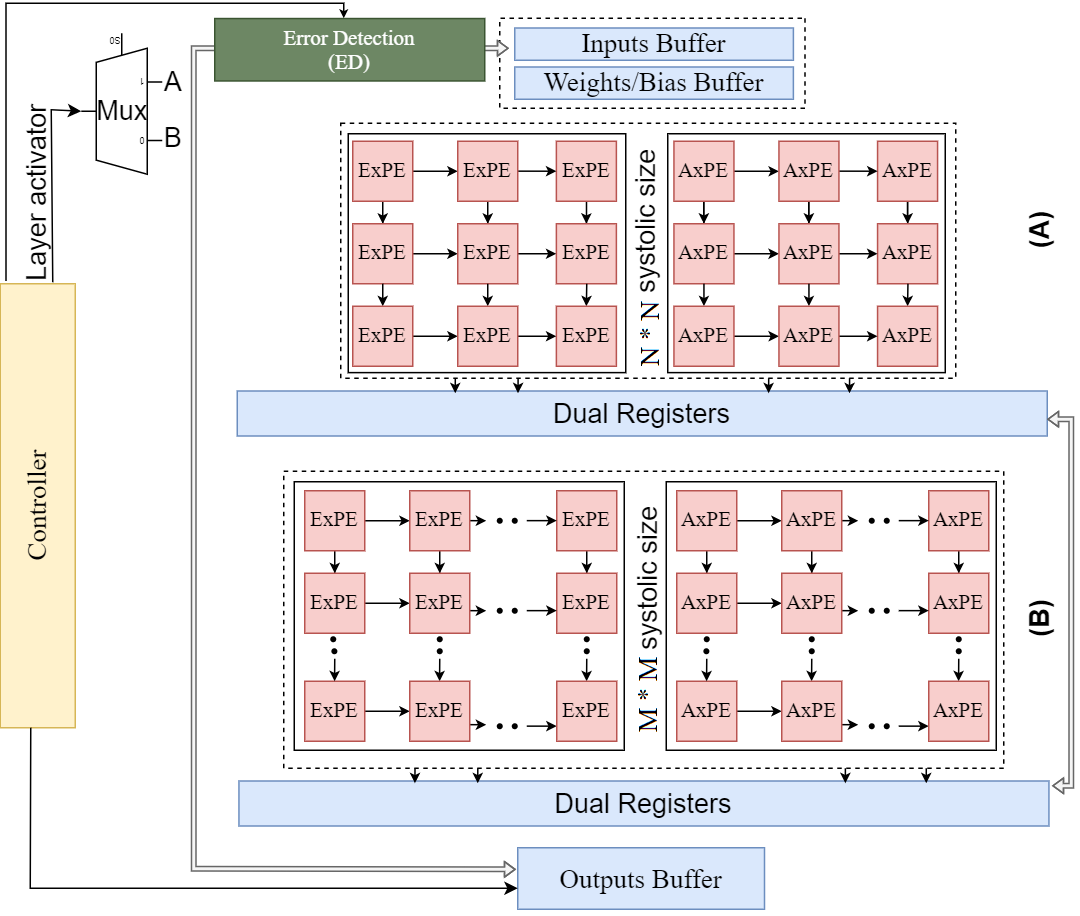}
    \caption{Proposed systolic architecture for our Resiliency assessment DNN accelerator framework}
    \label{fig:acc-model}
\end{figure}

Fig. \ref{fig:acc-model} illustrates the accelerator model to perform resilience analysis on FPGA. It consists of two different systolic architecture designs based on the network under test. The $N \times N$ systolic architecture is used based on the convolution layers' kernel size to perform the most optimum dot matrix. At the same time, all designs have ExPE and AxPE to perform the resilience analysis and benefits of a dual register to store the results of both approximate systolic and exact systolic for further comparisons. An Error Detector (ED) module is also provided to compute the error generated at each neuron's output compared to the exact output and can be used for the neuron's vulnerability evaluation.

This implementation provides us following features:
\begin{enumerate}[label=(\alph*)]
    \item Understanding the vulnerability of neurons by computing the error generated through the hardware and further layers by comparing the exact and approximate systolic design outputs;
    \item Increasing the controllability for enabling errors in each layer individually and keeping the other layers correct;
    \item Eliminating the need for designing and deploying an extra complex controller for the fault injection procedure. A simple approximate unit enabling circuitry is employed instead;
    \item The inference pipeline process executes a batch of inputs with no need to break this process;
    \item The resilience assessment process is performed without an extra interconnect for weight sampling;
    \item The proposed approach is not iterative for each potential fault location (unlike the traditional fault injection). Thus, the analysis complexity is vastly reduced.
\end{enumerate}
Note that the features (c)-(f) are specific for FI emulation in Programmable Logic and generally not available in Processing Logic based methods such as Fiji-FIN. 

\subsection{Experimental Results}
\subsubsection{Evaluation methodology}
To assess the feasibility of the proposed method, we implemented the same flow as shown in Fig. \ref{experiment} with fault injection (FI). Using Table I, we narrowed down the list of candidate approximate multipliers from the EvoApproxLib library \cite{evoapprox16} based on several relevant metrics, with a primary focus on two established features, namely, the Variance of Error Distance (Var-ED) and Root Mean Square (RMS-ED) presented in \cite{ansari2019improving}. These metrics are crucial in determining the approximation-induced errors that affect the performance of an AxC unit in DNNs. We selected mul8s\_1L2N for the experiment based on these metrics and results achieved from the high-level experiments on the network through the proposed GPU accelerated framework for CNN approximation in Section II.

For the reference part, we repeated the fault resiliency evaluation on the original network, which was instrumented with a state-of-the-art FI method \cite{fiji}. In this study, we considered the injection of multiple bitflips at a random location in all OFM' bits of the compromised layer for every input in the DNN validation test set. In this case, we assumed that 10\% of the weights' bits were faulty.

\begin{figure*}[h]
     \centering
     \begin{minipage}[b]{0.45\textwidth}
         \centering
         \includegraphics[width=1\textwidth]{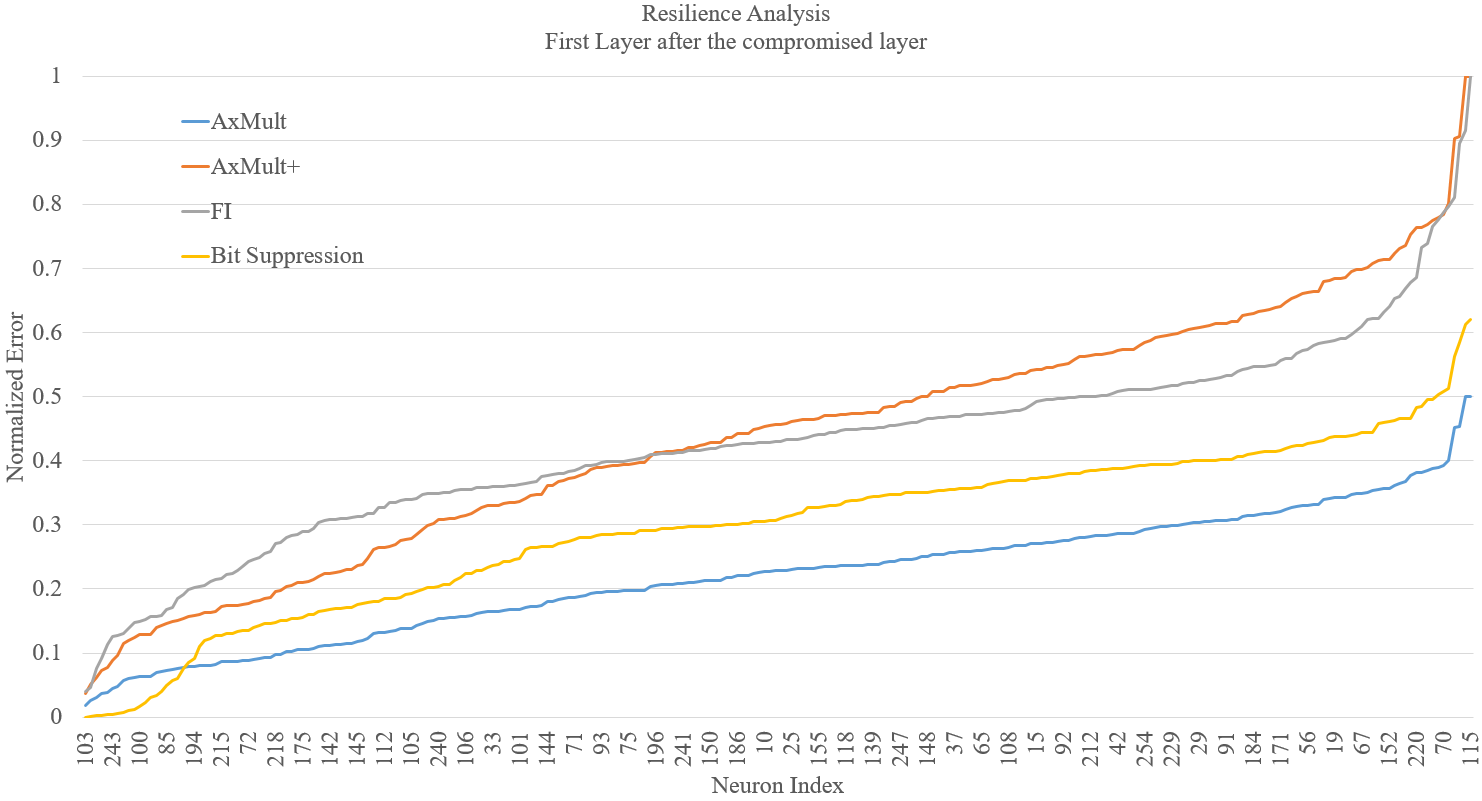}
         \caption{Normalized output error of Pool2: Applying AxMult, AxMult+ , Bit Suppression and FI on the Conv1}

         \label{res1}
     \end{minipage}
     \hfill
     \begin{minipage}[b]{0.5\textwidth}
         \centering
         \includegraphics[width=1\textwidth]{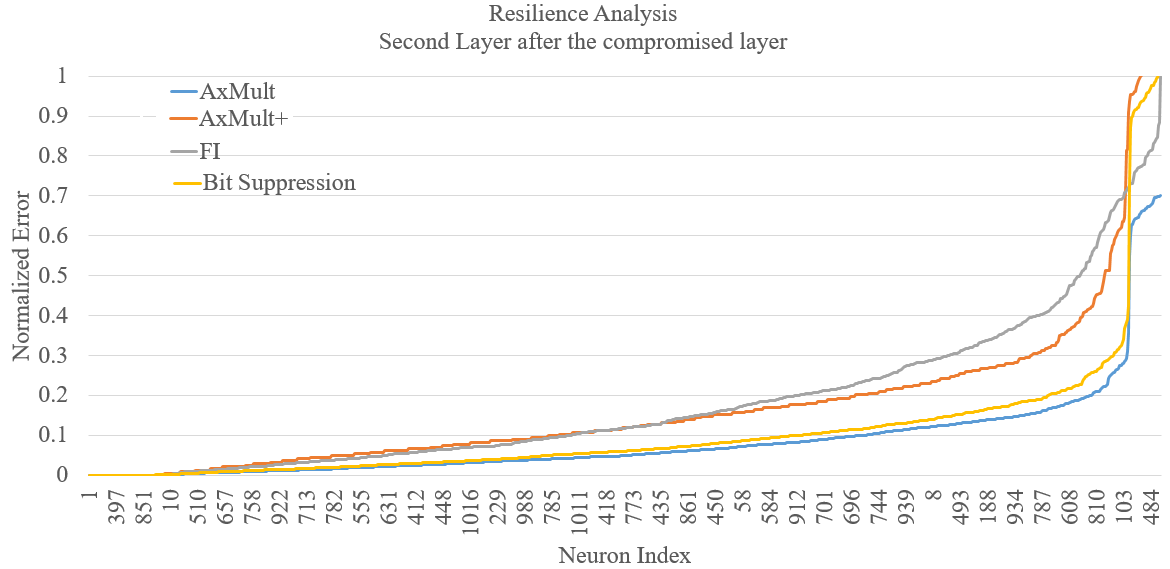}
         \caption{Normalized output error of Conv2: Applying AxMult, AxMult+ , Bit Suppression and FI on the Conv1}
         \label{res2}
     \end{minipage}
\end{figure*}

\begin{figure*}[h]
     \centering
     \begin{minipage}[b]{0.45\textwidth}
         \centering
         \includegraphics[width=1\textwidth]{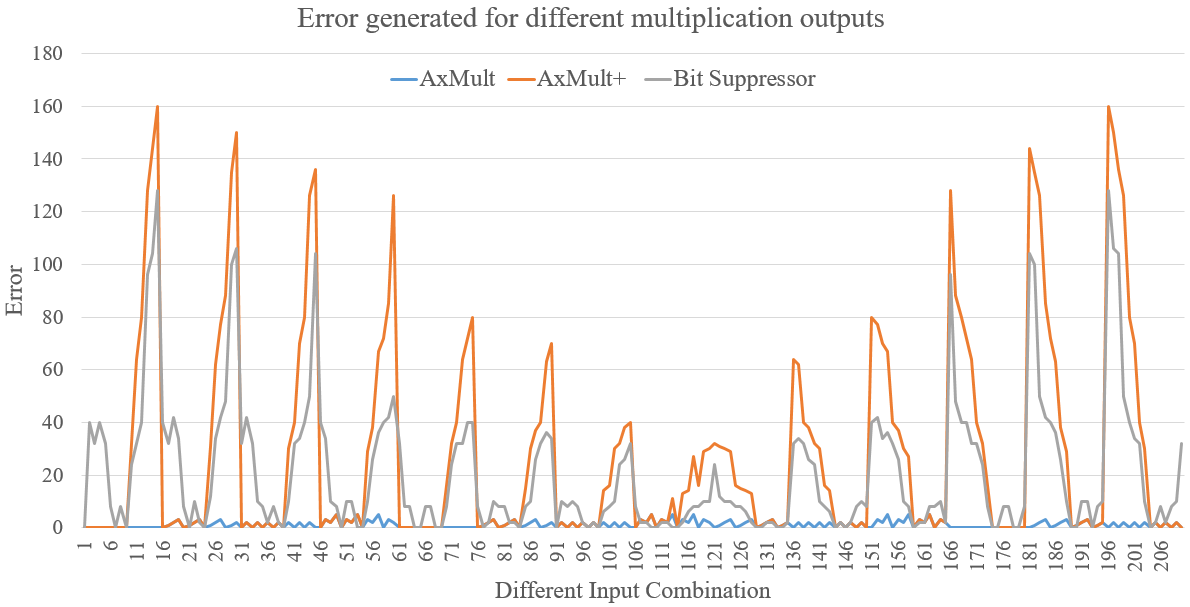}
         \caption{Multiplication output error generated by AxMult, AxMult+ and Bit Suppression}

         \label{res3}
     \end{minipage}
     \hfill
     \begin{minipage}[b]{0.45\textwidth}
         \centering
         \includegraphics[width=1\textwidth]{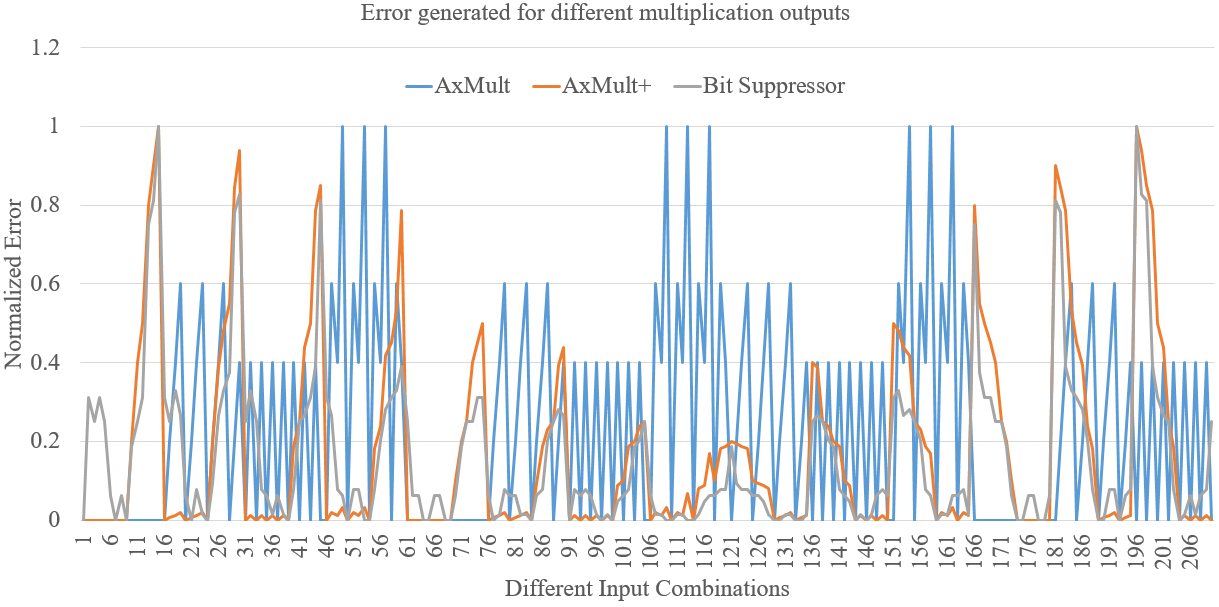}
         \caption{Normalized Multiplication output error generated by AxMult, AxMult+, and Bit Suppression}
         \label{res4}
     \end{minipage}
\end{figure*}
To achieve a high FI confidence level using the statistical fault injection approach \cite{leveugle2009statistical}, we repeated the experiment for each fault model with 1000 random faults per image. The average accuracy of all repetitions was then reported.

We evaluated the impact of AxMult, AxMult + Bit Suppression (AXMult+), Bit Suppression alone, and fault injection, along with normalized error and the number of flipped bits, on the DNN accuracy. The results show a drop in DNN accuracy due to these factors. We compared the normalized error and the number of flipped bits for each scenario.

\subsubsection{Experimental Setup}
To evaluate the feasibility of the proposed method, a case-study Convolutional Neural Network (CNN) with two convolutional layers, two max-pooling, and one Fully-Connected (FC) layer was implemented and trained. The simulations were performed on an Intel® Core™ i7-6800K CPU @ 3.40GHz × 12, and the proposed method was implemented with Python 3. The hardware synthesis and implementation results are produced by the Xilinx Vivado HLS tool on a Xilinx Versal VCK190 FPGA (xcvc1902-vsva2197-2MP-e-S) at 166 MHz operational frequency.

The CNN under study is trained on a dataset of 2000 images of animals (cats and dogs) and humans for binary classification. The accuracy of the network over the test set (including 450 images of animals and humans) is 93.34\%. Bit truncation quantization is applied in network parameters during training, and data precision is reduced to 8-bit. 
\subsubsection{Evaluation Results}

We analyzed the similarity of the fault resiliency analysis results obtained by fault emulation and our proposed method using the metrics identified in Section \ref{pr}.

Fig. \ref{res2} shows the distribution of \textit{normalized error} in the output of the second convolutional layer (Conv2) in the presence of 10\% random faults in the first convolution layer (grey), errors induced by AxMult (blue), and errors induced by AxMult + bit suppressor (orange) enabled in the first convolution layer, respectively. Fig. \ref{res1} reports the impact of applying FI and our proposed method on the same convolutional layer and its effect on the second pooling layer of the network. These results demonstrate the similarity in error propagation trends between the proposed and reference methods.

In practice, by analyzing these charts, users can set a criticality threshold on the output error of the neurons based on their application and determine the number and indices of neurons to be used for any protection techniques. Generally, if we set the threshold at some error value, all methods suggest some neuron indices for mitigation techniques. As it can be concluded, both AxMult + bit suppression and FI show very similar behaviors. However, relying solely on the AxMult or bit suppression techniques is quite inaccurate for high fault ratios like this case study here.

For example, by setting the error threshold to 0.7, FI will recommend the user to protect 50 out of 1024 neurons of the Conv2 network's second CONV's neurons, while AxMult + bit suppression will recommend 53 out of 1024 neurons, including all the critical neurons recognized by FI. Fig. \ref{res3} and Fig. \ref{res4} show the error distribution of the three different methods, i.e., AxMult, AxMult + bit suppressor, and bit suppressor on the output of a multiplication operation with all the combinations of two 8-bit inputs. From Fig. \ref{res3}, it is evident that the error values generated by AxMult + bit suppressor can almost cover a vast range of different values, and Fig. \ref{res4} shows that the error is evenly distributed on all different input combinations.

Table. \ref{bitflip} is reporting the number of bitflips and accuracy drop in subsequent layers caused by the compromised first convolution layer. These results also demonstrate the strong similarity of the trends in error propagation by the AxMult and its variants with the reference method. In case of accuracy drop, AxMult + bit suppression shows a strong correlation with the FI method and surpasses the other two methods.

Table \ref{hard1} reports details of the hardware accelerator implementation. Based on the results, the proposed implementation can be executed on the FPGA at 166 MHz clock frequency, and only by using $\sim$16\% of the available LUTs on the board all three mentioned systolic-array size architectures can be implemented to improve the efficiency of the accelerator. The timing comparison of the proposed method and the state-of-the-art fault injection method are presented in Table. \ref{hard}. As it can be concluded, by keeping an acceptable accuracy of FI in identifying the critical neurons, we get thousands of times speed-up in the resilience assessment of the DNNs. (Specifically, it is 5417 times in this example). At the same time, the proposed method does not need extra interconnects to manage the assessment process, and the original controller of the accelerator can take care of the fault resiliency assessment process.

\begin{table}[]
\caption{Bitflips and Accuracy drop induced by our proposed method vs. the reference fault injection method by fault rate 10\% in OFM of the first convolution layer}

\label{bitflip}
\resizebox{\columnwidth}{!}{%
\begin{tabular}{|ccccc|}
\hline
\multicolumn{1}{|c|}{\multirow{3}{*}{\textbf{Measured  Layer}}} & \multicolumn{4}{c|}{\textbf{Bitflips in subsequent layers}}                                                                                                                                                                                                                                                                                              \\ \cline{2-5} 
\multicolumn{1}{|c|}{}                                 & \multicolumn{1}{c|}{\begin{tabular}[c]{@{}c@{}}\textbf{FI (reference)}\\ {[}\%{]}\end{tabular}} & \multicolumn{1}{c|}{\begin{tabular}[c]{@{}c@{}}\textbf{AxMult}\\ {[}\%{]}\end{tabular}} & \multicolumn{1}{c|}{\begin{tabular}[c]{@{}c@{}}\textbf{AxMult+} \\ {[}\%{]}\end{tabular}} & \begin{tabular}[c]{@{}c@{}}\textbf{Bit suppressor}\\ {[}\%{]}\end{tabular} \\ \hline
\multicolumn{1}{|c|}{Conv1}                            & \multicolumn{1}{c|}{10}                                                                & \multicolumn{1}{c|}{10.30}                                                     & \multicolumn{1}{c|}{10}                                                                           & 10.20                                                             \\ \hline
\multicolumn{1}{|c|}{Pool1}                            & \multicolumn{1}{c|}{9.07}                                                              & \multicolumn{1}{c|}{9.20}                                                      & \multicolumn{1}{c|}{9.06}                                                                         & 9.15                                                              \\ \hline
\multicolumn{1}{|c|}{Conv2}                            & \multicolumn{1}{c|}{16.76}                                                             & \multicolumn{1}{c|}{16.80}                                                     & \multicolumn{1}{c|}{16.77}                                                                        & 16.83                                                             \\ \hline
\multicolumn{1}{|c|}{Pool2}                            & \multicolumn{1}{c|}{16.51}                                                             & \multicolumn{1}{c|}{16.66}                                                     & \multicolumn{1}{c|}{16.53}                                                                        & 16.62                                                             \\ \hline
\multicolumn{5}{|c|}{\textbf{Accuracy drop} {[}\%{]}}                                                                                                                                                                                                                                                                                                                                                             \\ \hline
\multicolumn{1}{|c|}{}                                 & \multicolumn{1}{c|}{16.73}                                                             & \multicolumn{1}{c|}{9.33}                                                      & \multicolumn{1}{c|}{18.33}                                                                        & 24.73                                                             \\ \hline
\end{tabular}}
\end{table}

\begin{table}[]
\caption{Hardware implementation of the proposed hardware accelerator}
\resizebox{\columnwidth}{!}{%
\label{hard1}
\begin{tabular}{|c|ccc|c|c|}
\hline
                                                               & \multicolumn{3}{c|}{\textbf{Resource Utilization (\%)}}                &                                                                                                &                                                         \\ \hline
\begin{tabular}[c]{@{}c@{}}\textbf{Conv2D}\\ \textbf{systolic size}\end{tabular} & \multicolumn{1}{c|}{\textbf{LUT}}   & \multicolumn{1}{c|}{\textbf{FF}}   & \textbf{BRAM} & \begin{tabular}[c]{@{}c@{}}\textbf{Data Path}\\ \textbf{Delay}\end{tabular}                                      & \begin{tabular}[c]{@{}c@{}}\textbf{CLK}\\ \textbf{Frequency}\end{tabular} \\ \hline
3*3                                                            & \multicolumn{1}{c|}{0.03}  & \multicolumn{1}{c|}{0.00} & 0.83 & \multirow{3}{*}{\begin{tabular}[c]{@{}c@{}}Logic: $\sim$20\%\\ Route: $\sim$80\%\end{tabular}} & \multirow{3}{*}{166 MHz}                                \\ \cline{1-4}
5*5                                                            & \multicolumn{1}{c|}{0.09}  & \multicolumn{1}{c|}{0.00} & 0.83 &                                                                                                &                                                         \\ \cline{1-4}
32*32                                                          & \multicolumn{1}{c|}{15.30} & \multicolumn{1}{c|}{0.91} & 0.85 &                                                                                                &                                                         \\ \hline
\end{tabular}}
\end{table}


\begin{table}[h!]
\caption{Timing overheads of the proposed method vs. the reference fault injection method (Conv1 layer)}
\label{hard}
\tiny
\resizebox{.5\textwidth}{!}{
\centering
\begin{tabular}{|cccc|}
\hline
\multicolumn{1}{|c|}{\textbf{Network}}                                                                    & \multicolumn{1}{c|}{\begin{tabular}[c]{@{}c@{}}\textbf{Analysis Control} \\ \textbf{Circuitry}\end{tabular}}     & \multicolumn{1}{c|}{\textbf{Interconnects}}                                                                  & \begin{tabular}[c]{@{}c@{}}\textbf{DNN execution}\\ \textbf{time in FPGA}\end{tabular} \\ \hline
\multicolumn{1}{|c|}{Base CNN}                                                                   & \multicolumn{1}{c|}{N/A}                                                                      & \multicolumn{1}{c|}{\begin{tabular}[c]{@{}c@{}}Data Exchange \\ Interconnect\end{tabular}}          & $\sim$120ms                                                            \\ \hline
\multicolumn{4}{|c|}{\textbf{Fault Resilience Assessment}}                                                                                                                                                                                                                                                                                                                                                                                                                   \\ \hline
\multicolumn{1}{|c|}{\begin{tabular}[c]{@{}c@{}}CNN instrumented \\ with FI\end{tabular}}        &  \multicolumn{1}{c|}{Complex FI Controller}                                                    & \multicolumn{1}{c|}{\begin{tabular}[c]{@{}c@{}}(Data Exchange + FI) \\ Interconnect\end{tabular}}   & $\sim$650,000ms                                                            \\ \hline
\multicolumn{1}{|c|}{\begin{tabular}[c]{@{}c@{}}CNN instrumented\\  with AxMult+\end{tabular}} & \multicolumn{1}{c|}{\begin{tabular}[c]{@{}c@{}}Accelerator \\ Controller\end{tabular}} & \multicolumn{1}{c|}{\begin{tabular}[c]{@{}c@{}}Data Exchange \\ Interconnect\end{tabular}} & $\sim$120ms                                                   \\ \hline
\end{tabular}}
\end{table}

\section{Fault resiliency in DNNs} \label{sec:fault-mask}

\subsection{Motivations and Related Works}
In the last few years, researchers have investigated the theory behind brain-inspired computational models to build artificial structures capable of addressing highly complex computational problems. Today, DNNs are considered attractive solutions in several fields due to their outstanding computational capabilities as well as their human-level performance. The human brain is a complex and fascinating system able to bear synapses or neuron faults and still keep working properly, thanks to its plastic ability to remodel, repair, and reorganize its neural functions. Similarly, artificial neural networks possess in their structure a certain degree of redundancy that leads to intrinsic robustness and resilience against the occurrence of faults. This is caused by two aspects: the first is related to their distributed and parallel structure; the second to the redundancy resulting from the over-provisioning \cite{PIURI200118}. Indeed, neural networks are furnished with a quantity of artificial neurons higher than the minimal number required to perform a computation. It means that they can bear a bounded number of errors thanks to the excessive neuron budget: once this number is exceeded, the precision degrades gracefully as the number of errors increases \cite{when}. 

This structural feature allows them to have an attractive property known as \textit{masking ability}, which corresponds to the ability of DNNs to stop the propagation of some faults by masking their effects. As an example, it has been shown that the presence in DNNs of the Rectified Linear Unit (ReLU) activation function halves the percentage of critical faults by stopping the propagation of faults on negative weights \cite{etsRELU}. Understanding how faults propagate through the neural network is very important, as it may influence: the reliability assessment procedure; efficient fault detection and mitigation strategies. 

The analysis of fault propagation in DNNs has been conducted in the literature by different perspectives. 
A preliminary theory-driven analysis is proposed in \cite{formulePropagation}, where the authors explore inherent characteristics of fault propagations in DNNs from the theoretical aspect.
They propose a formula to compute the perturbation caused by the \textit{i-}th bit flip on a weight represented in a 32-bit floating-point format. The authors in \cite{UnderstandingNVIDIA} characterize the propagation of soft errors from the hardware to the application software of \acrshort{dnn} systems. Based on this, they devise cost-effective solutions to mitigate Silent Data Corruption (SDC) in software and hardware. Further studies on faults propagations in DNNs are described in \cite{itc22Esteban} and \cite{rechPropagation}.

Nevertheless, it is important to underline that the major effort in the above-mentioned research works consists in understanding how critical faults (i.e., those that lead to application failures) propagate through the hardware-software system. 

The intent of this section is twofold. On the one hand, it aims to show how a critical fault spreads through a network. On the other hand, this section tackles the problem from a different angle, showing how a \textit{masked fault} is propagated within the system, analysing the role DNNs have in this process. The investigation of this latter category of faults is important for the following reasons: 
\begin{itemize}
    \item In a fault injection process, the identification of sets of faults that are masked may reduce the fault space and, as a consequence, lower the costs of the reliability assessment; 
    \item In the design of \acrshort{dnn} models, the knowledge of architectural elements that favour the masking ability of DNNs can lead to the design of more robust models.
\end{itemize}
This section presents an analysis on masked faults with the goal of identifying at what point in the computation their propagation is stopped and if it is possible to know in advance their effects on the output of the DNN. 

\subsection{Proposed Method}
\label{sec:proposed}

\begin{figure*}[!th]

\centering

\begin{subfigure}{.33\textwidth}
  \centering
  \includegraphics[width=.99\linewidth]{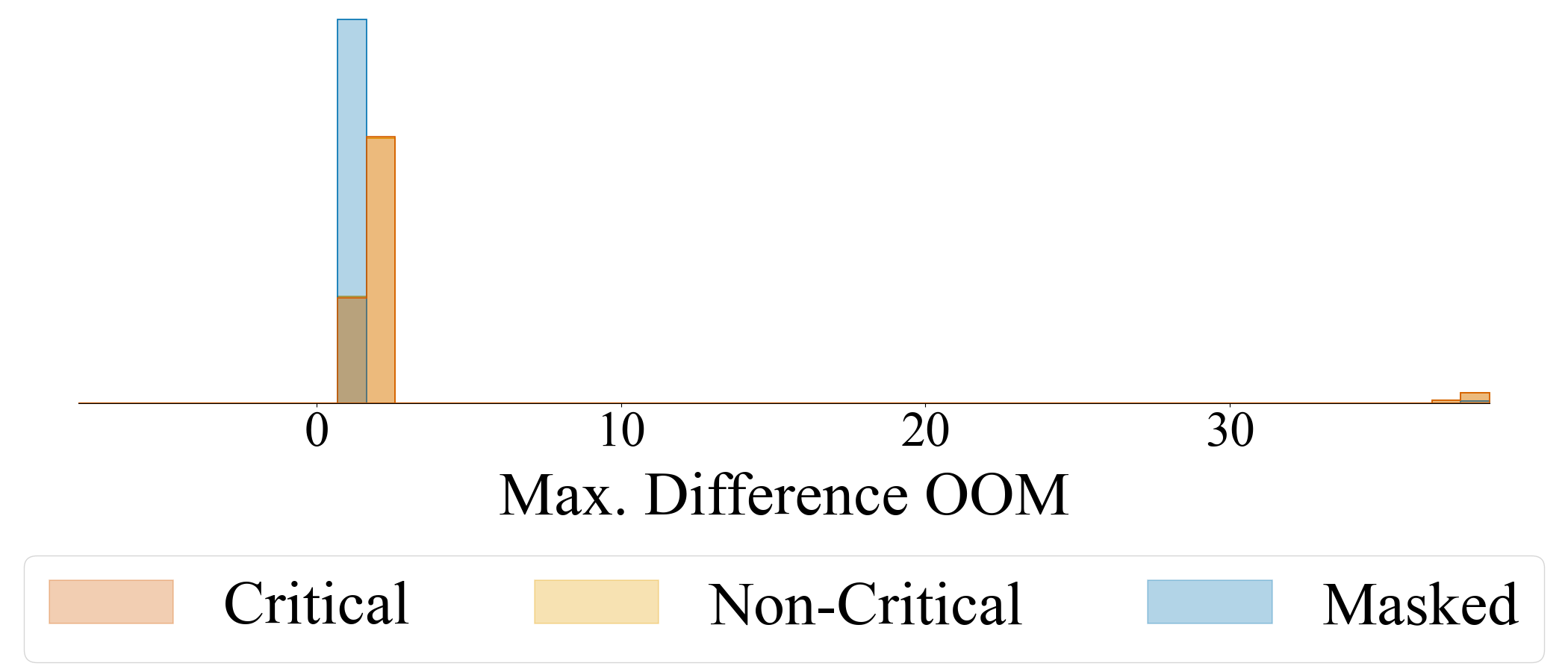}
  \caption{LeNet5: Max. Difference}
  \label{fig:lenet_maxdiff}
\end{subfigure}%
\begin{subfigure}{.33\textwidth}
  \centering
  \includegraphics[width=.99\linewidth]{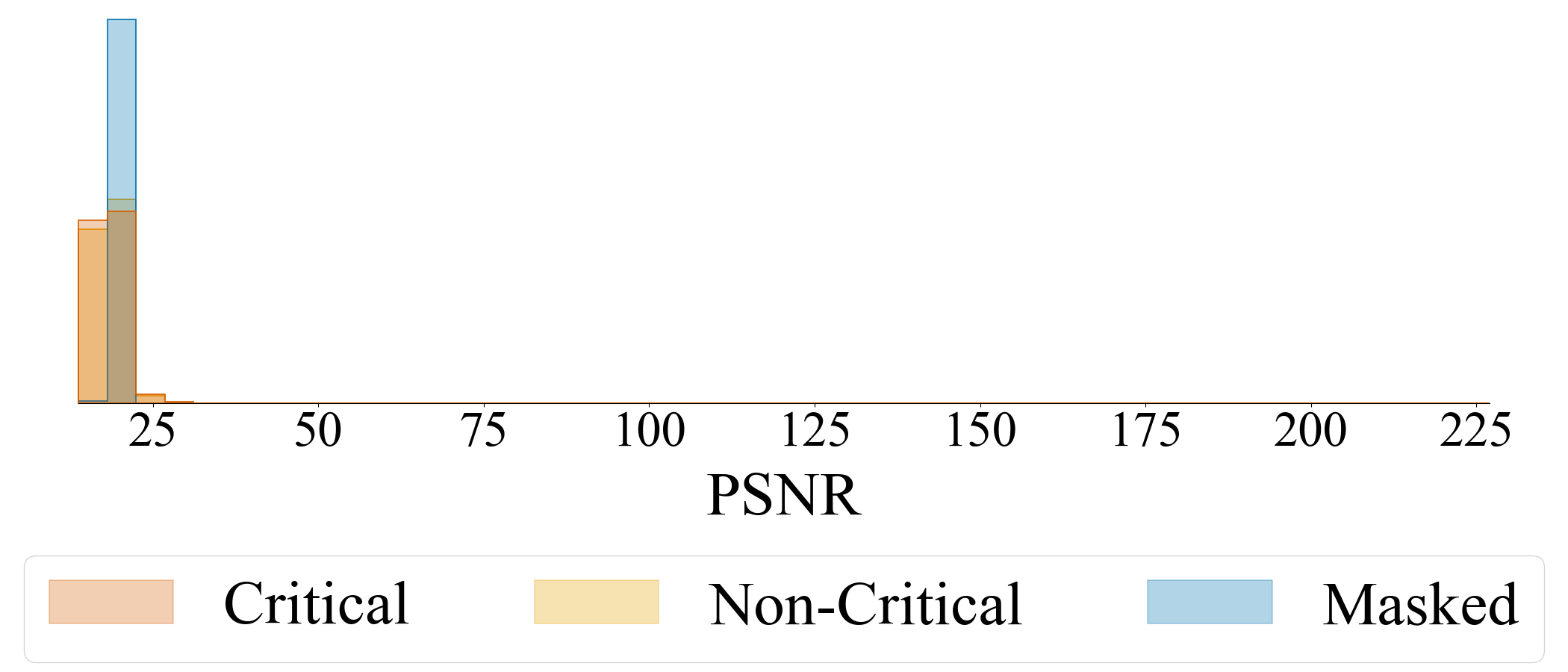}
  \caption{LeNet5: PSNR}
  \label{fig:lenet_PSNR}
\end{subfigure}
\begin{subfigure}{.3\textwidth}
  \centering
  \includegraphics[width=.99\linewidth]{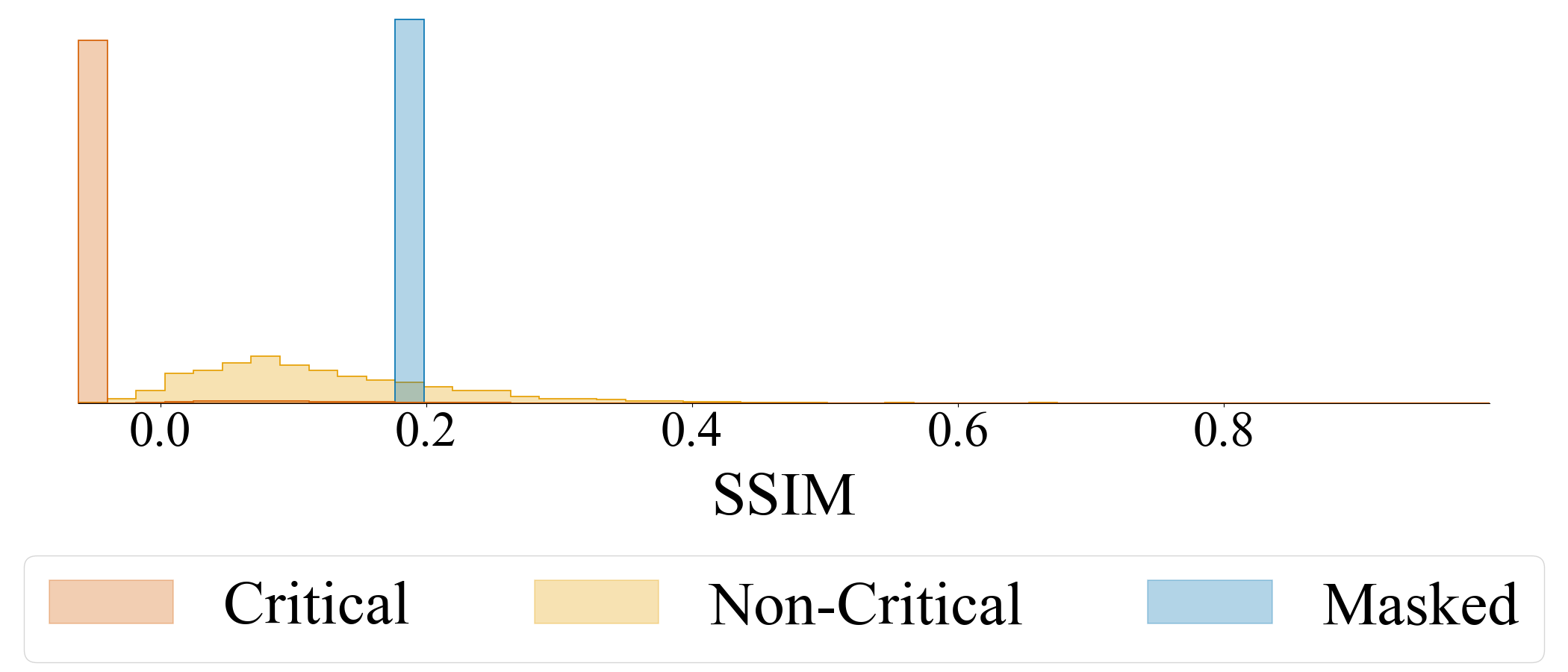}
  \caption{LeNet5: SSIM}
  \label{fig:lenet_SSIM}
\end{subfigure}

\begin{subfigure}{.33\textwidth}
  \centering
  \includegraphics[width=.99\linewidth]{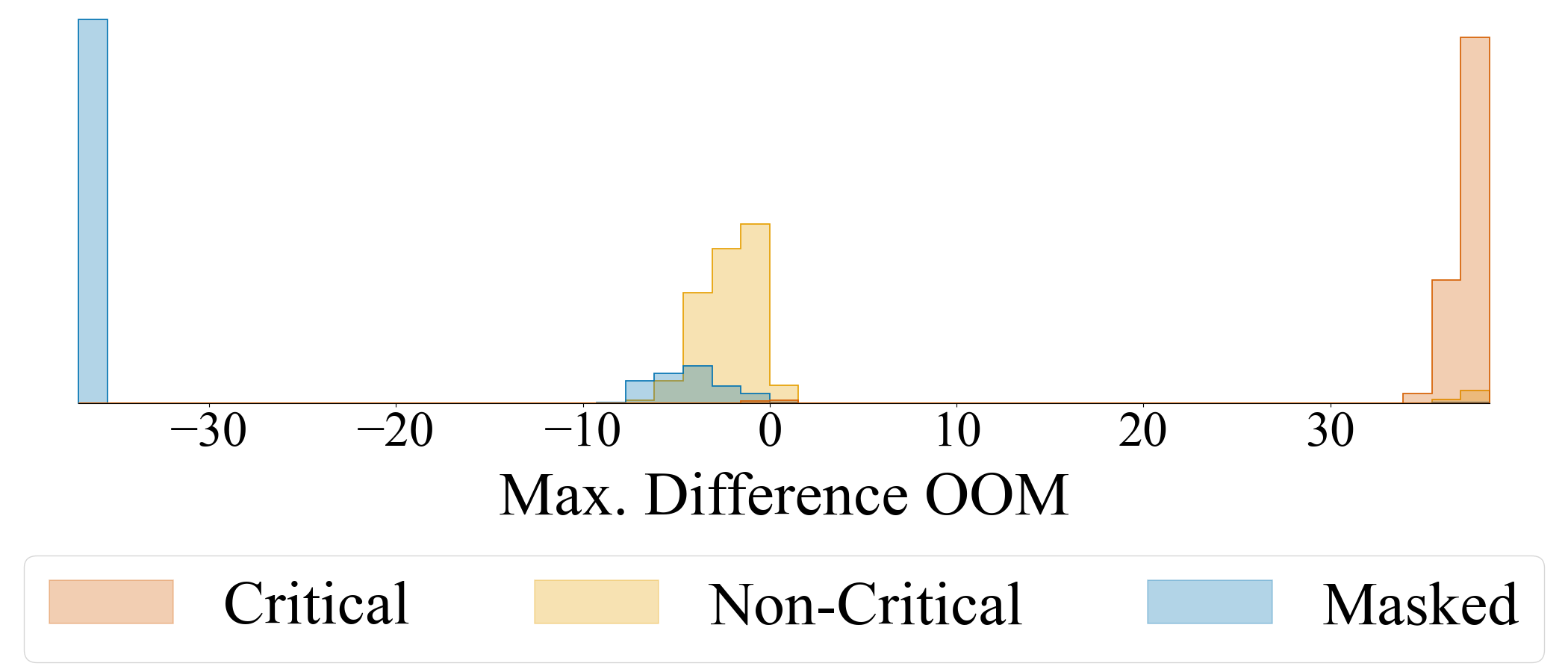}
  \caption{ResNet20: Max. Difference}
  \label{fig:resnet_maxdiff}
\end{subfigure}%
\begin{subfigure}{.33\textwidth}
  \centering
  \includegraphics[width=.99\linewidth]{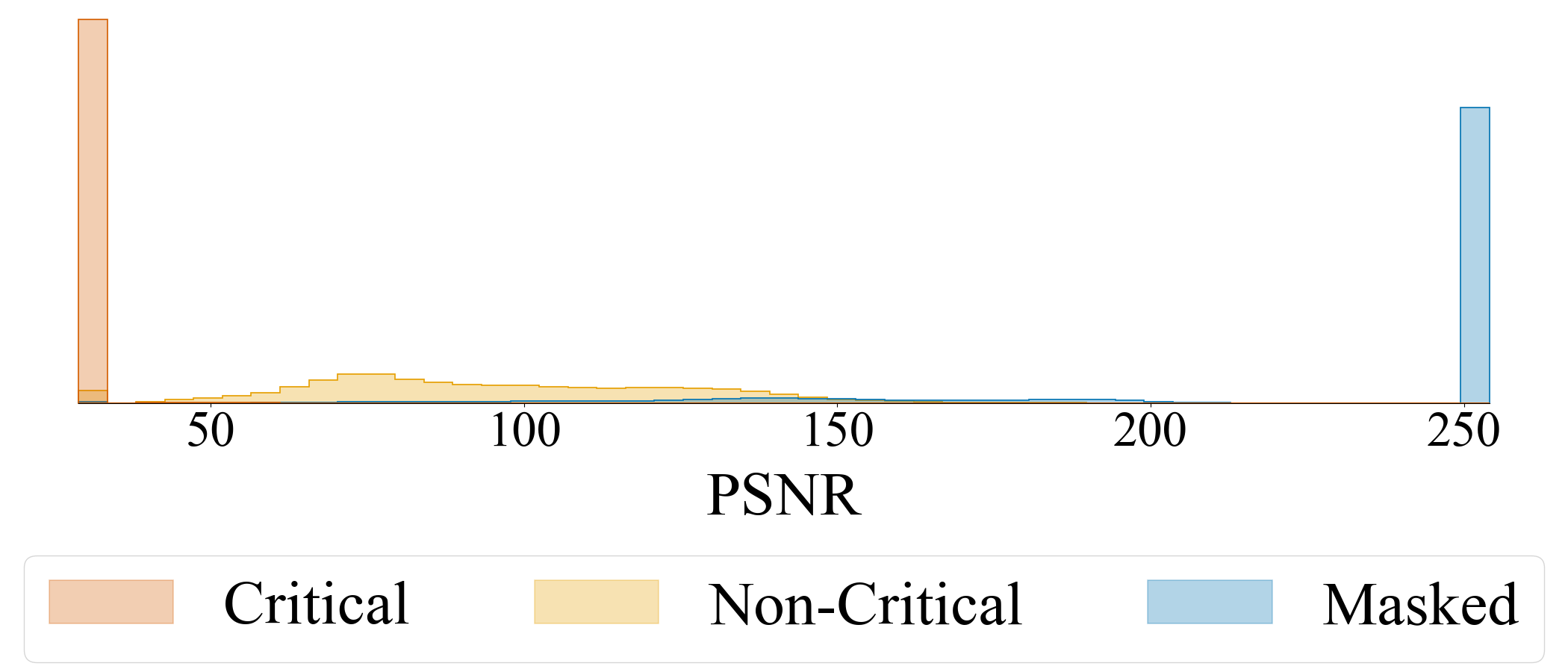}
  \caption{ResNet20: PSNR}
  \label{fig:resnet_PSNR}
\end{subfigure}
\begin{subfigure}{.3\textwidth}
  \centering
  \includegraphics[width=.99\linewidth]{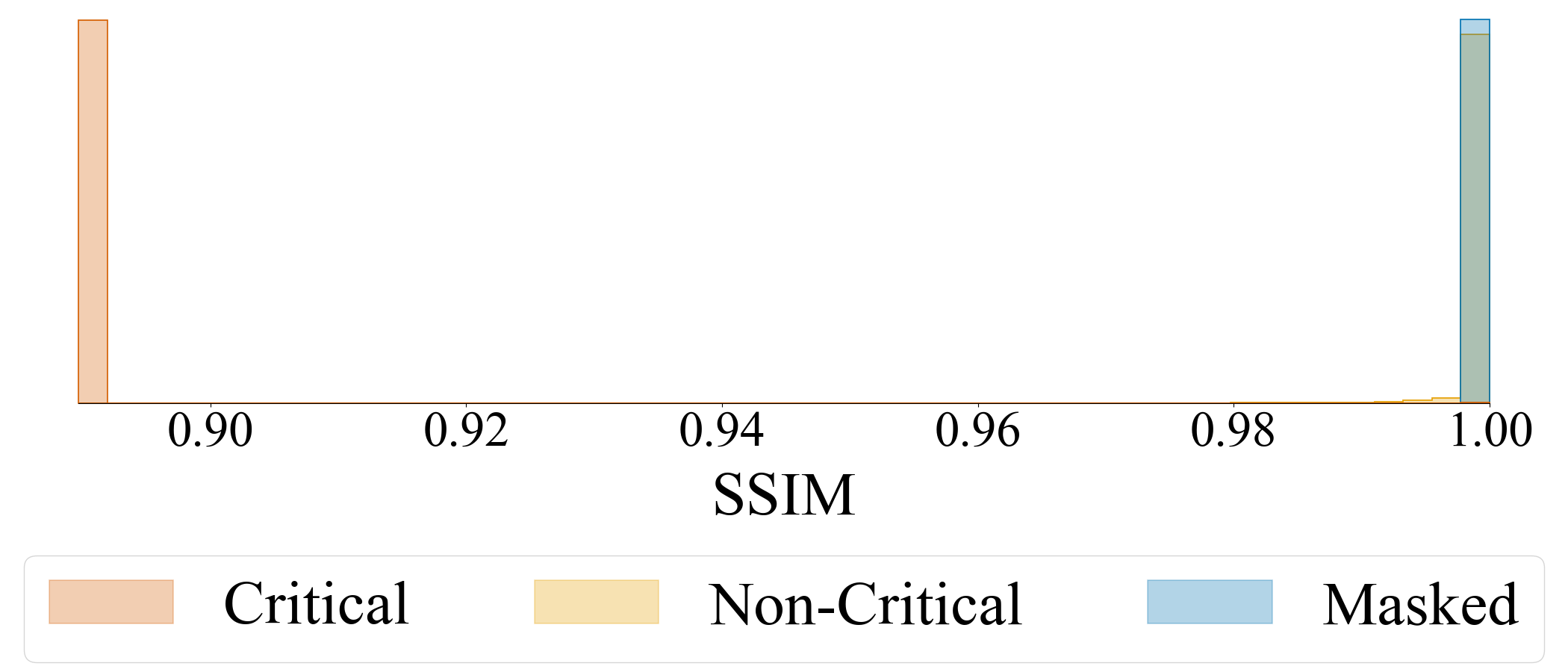}
  \caption{ResNet20: SSIM}
  \label{fig:resnet_SSIM}
\end{subfigure}

\begin{subfigure}{.33\textwidth}
  \centering
  \includegraphics[width=.99\linewidth]{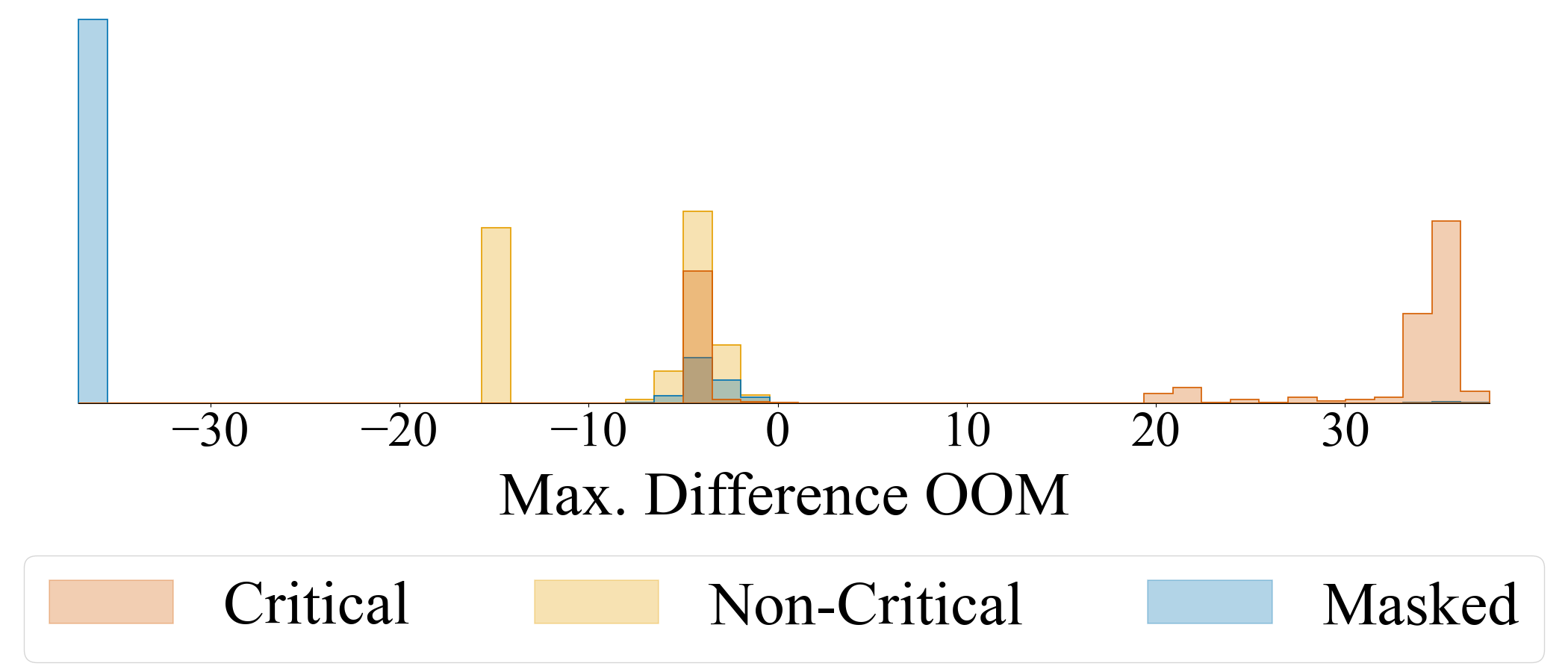}
  \caption{DenseNet121: Max. Difference}
  \label{fig:densenet_maxdiff}
\end{subfigure}%
\begin{subfigure}{.33\textwidth}
  \centering
  \includegraphics[width=.99\linewidth]{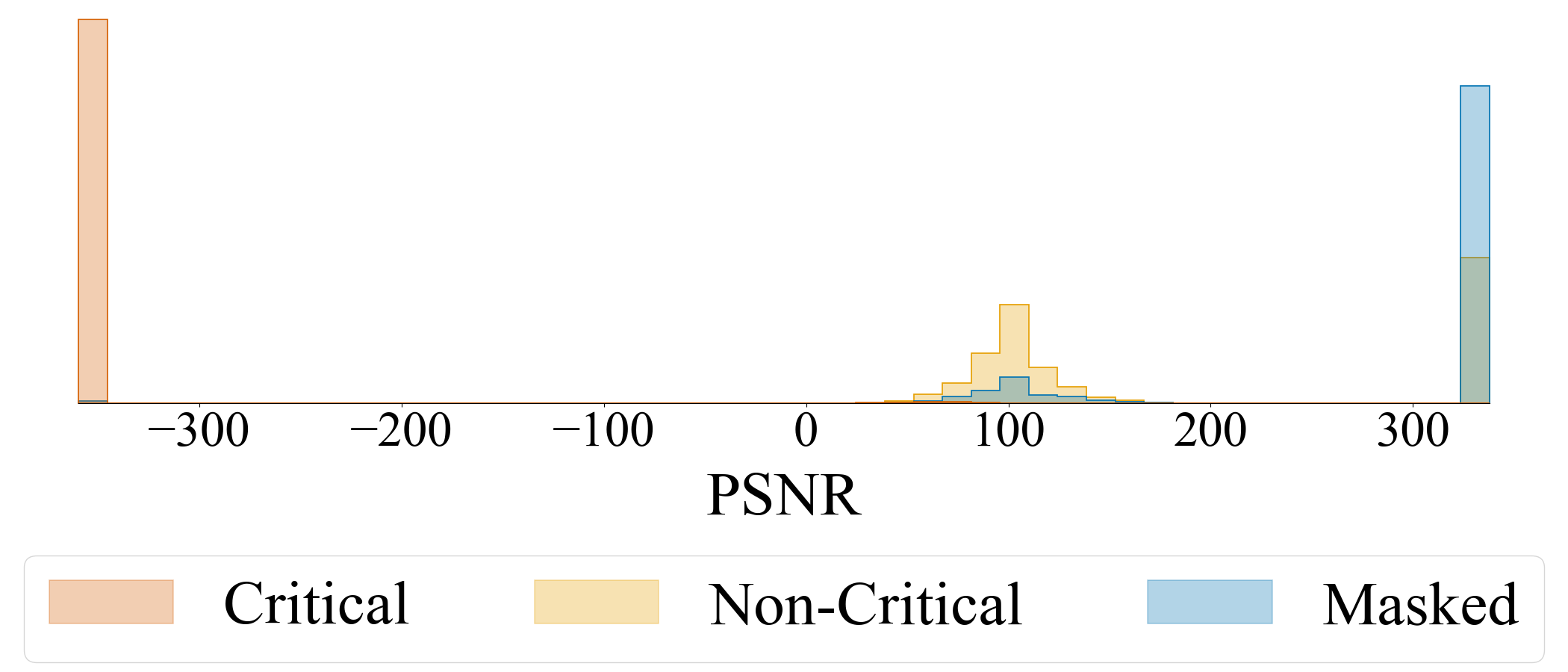}
  \caption{DenseNet121: PSNR}
  \label{fig:densenet_PSNR}
\end{subfigure}
\begin{subfigure}{.3\textwidth}
  \centering
  \includegraphics[width=.99\linewidth]{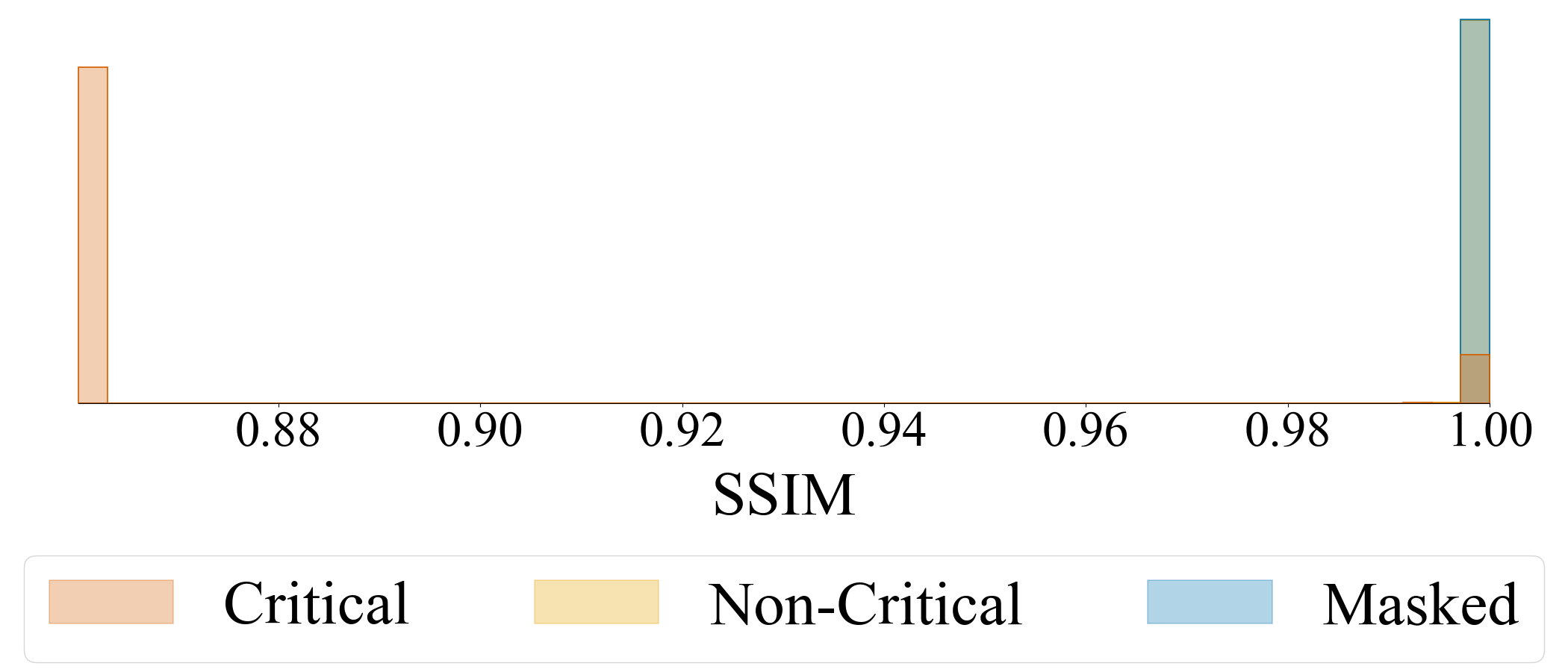}
  \caption{DenseNet121: SSIM}
  \label{fig:densenet_SSIM}
\end{subfigure}

\caption{Metric probability distributions for the CNN under exam, observed in the layer where the fault is injected. Figures (a)-(c) refer to LeNet-5, Figures (d)-(f) refer to ResNet20 and Figures (g)-(i) refer to DenseNet121.}
\label{fig:dist}
\end{figure*}

\glspl{cnn} are a subset of \glspl{dnn} composed of a set of convolutional layers. The output of each layer is a multidimensional tensor, often referred to as the \textit{Output Feature Map} (OFM). In the field of Image Classification, the output of the network is represented by a vector called \textit{logit}. A fault affecting a \acrshort{cnn} can be classified as:
\begin{itemize}
    \item \textbf{Critical}, if it causes a change in the network prediction;
    \item \textbf{Non-Critical}, if it impacts the logit without changing the prediction;
    \item \textbf{Masked}, if it does not modify the logit.

\end{itemize}
When a fault affects the parameters of a layer (i.e., weights), it may change its OFM, as well as the one of all the following layers. If the fault is masked, the difference between the \textit{golden Output Feature Map} (gOFM) and the \textit{faulty Output Feature Map} (fOFM) of the impacted layer should be small or zero. Contrarily, it is logical to assume that a critical fault also produces a fOFM that is radically different from the gOFM.

As a consequence of these two observations, it is possible to predict the impact of a fault without needing to carry out a complete inference. In fact, this section aims at showing that:
\begin{enumerate}
    \item Masked faults, once triggered, rarely propagate for more than one layer. Thus, the only different OFM is the one of the layer directly affected by the fault;
    \item Critical faults, can be immediately identified by performing some early measures, using some metrics that can be computed by comparing the fOFM and the gOFM of the affected layer. 
\end{enumerate}

The OFM of a layer $l$ can be interpreted as a collection of $n$ filtered images, where $n$ is the number of filters applied in layer $l$. Furthermore, the fOFM resulting from a fault in the network parameters can be interpreted as the gOFM plus a Gaussian noise. 
Therefore, it is possible to apply well-known objective image quality metrics, such as the Peak signal-to-noise Ratio (PSNR) and the Structural Similarity Index Metric (SSIM) \cite{5596999}. 

This section proposes to use three different metrics to predict the criticality of a fault, starting from the OFM of the affected layer.

\subsubsection{Max Difference} This first metric computes the maximum distance between the gOFM and the fOFM. This metric is presented as a baseline since, to the best of the authors' knowledge, there are no metrics that correlate the criticality of a fault with the changes in the OFM.

\subsubsection{PSNR} This metric is directly proportional to the ratio between the peak signal (i.e., the maximum element of the gOFM) and the power of the corrupting noise, represented by the mean square error between the gOFM and the fOFM. The value can be computed as follows:

\begin{equation}
    PSNR = 10 \cdot \log_{10} \frac{\max(gOFM)^2}{MSE(gOFM, fOFM)} 
\end{equation}

Where $\max(gOFM)$ is the maximum value of the gOFM and $MSE$ is the Mean Square Error between the gOFM and the fOFM. 

\subsubsection{SSIM} this metric improves the PSRN, by including the concept of \textit{structural information}, represented by the relationship of a neuron with its neighbours. The formula is composed by the product of three terms, the \textit{luminance}, the \textit{contrast} and the \textit{structural} term. In the context of the study of the OFM, the simplified formula can be expressed as:

\begin{equation}
    SSIM = \frac{(2\mu_f\mu_g + C_1)(2\sigma_{fg} + C_2)}{(\mu_f^2 + \mu_g^2 + C_1)(\sigma_f^2 + \sigma_g^2 + C_2)}
\end{equation}

Where $\mu_g, \mu_f$ are the mean of the gOFM and of the fOFM, $\sigma_g, \sigma_f$ their standard deviation, $\sigma_{fg}$ their cross-covariance. $C_1$ and $C_2$ are two regularization parameters.

\subsection{Experimental Results}
This section analyses three different \glspl{cnn} used for Image Classification to study how a fault can propagate. The networks under analysis are: LeNet-5 with the MNIST dataset, ResNet20 with CIFAR-10 and Densenet-121 with ImageNet. For each network, we performed a statistical \acrshort{fi} as described in \cite{Date23}. The tool used to carry out the \acrshort{fi} campaign is SCI-FI \cite{SCI-FI}, that allows to speed up the \acrshort{fi} process using the Fault Dropping and the Delayed Start techniques. The faults injected are single bit-flips in the network parameters, represented as 32-bit floating points. Further details on the networks under exam and on the \acrshort{fi} campaigns are reported in Table \ref{tab:net_desc}.

\begin{table}[t]
\caption{The networks under analysis}
\centering
\begin{tabular}{|llrrrr|}
\hline
\textbf{Network} &
  \textbf{Dataset} &
  \multicolumn{1}{c}{\textbf{\begin{tabular}[c]{@{}c@{}}Dataset\\ Size\end{tabular}}} &
  \multicolumn{1}{c}{\textbf{\begin{tabular}[c]{@{}c@{}}Acc.\\ {[}\%{]}\end{tabular}}} &
  \multicolumn{1}{l}{\textbf{Weights}} &
  \multicolumn{1}{c|}{\textbf{\begin{tabular}[c]{@{}c@{}}Injected\\ Faults\end{tabular}}} \\ \hline
LeNet5      & MNIST    & 10,000 & 98.85 &    61,706 &  2,212 \\
ResNet20    & CIFAR-10 & 10,000 & 91.72 &   269,722 & 15,675 \\
DenseNet121 & ImageNet & 50,000 & 74.43 & 7,978,856 & 16,685 \\ \hline
\end{tabular}
\label{tab:net_desc}
\vspace{10pt}
\end{table}

Firstly, to demonstrate that Masked faults only modify the OFM of the layer affected by the fault, we report the percentage of Masked faults that affect more than one layer. In particular, for LeNet5, all the Masked faults do not modify any OFM besides the one of the impacted layer. For ResNet20, $87.99\%$ of Masked faults show no effect in the OFM of the layer immediately after the impacted one, while for DenseNet this number rises to $99.17\%$.

To show that Critical faults have a strong impact early on, we compute the metrics introduced in Section \ref{sec:proposed} on the OFM of the layer affected by the fault. Figure \ref{fig:dist} reports the metrics distributions for the Max Difference, the PSNR and the SSIM. Each image shows, for each network, the distribution of a metric computed for all the \acrshort{fi} campaigns. In particular, the distribution is further subdivided according to the impact of the fault affecting the network when they were measured. This means that the distribution labelled 'Critical' reports only the value measured when a Critical fault is affecting the network. For a metric, the more separable the three distributions are, the better the metric is at predicting the effect of a fault.

In particular, we can observe a stark contrast between the metrics computed for LeNet5 and the other networks. This can be imputed to the lack of batch-normalization layers, that normalize the value of the weights (and of the OFM) between $[-1, 1]$. Consequently, even a 
bit-flip in the mantissa bits of the weight can have a large impact. Nonetheless, SSIM performs sufficiently well, as it correctly separates Critical and Non-Critical faults. 

For the other two CNNs, we can notice that both the Max. Difference and the PSNR separate Masked faults from Critical and Non-Critical faults. However, for ResNet20, SSIM outperforms the other metrics, as it completely splits ups Critical and Non-Critical faults while providing a good degree of separation between Masked and Non-Critical faults. Contrarily, for DenseNet-121, SSIM does not completely separate Masked from Critical. For this latter network, the best solution is offered by the PSNR.

Therefore, we observe how different metrics can correctly predict Masked and Critical faults without the need for a complete inference, by simply analysing the fOFM of the layer affected by the fault.

As a final note, we want to highlight that the cost of the computation of the metric is quite small, requiring only a portion of the time required for the computation of a whole layer. On average, the per-layer overhead added by the computation of one of the metrics is $76.51\%$ for LeNet5 $74.28\%$ for ResNet20 and $73.54\%$ for DenseNet121.

\section{Conclusions} \label{sec:conclusion}
This paper explored approximation and fault resiliency of \acrshort{dnn} accelerators. To allow fast evaluation of \acrshort{axc} \acrshort{dnn}, an efficient \acrshort{gpu}-based simulation framework was developed. The paper proposed a method for employing approximate (AxC) arithmetic circuits to agilely emulate errors in hardware without performing fault injection on the DNN.  Finally, it presented a fine-grain analysis of fault resiliency by examining fault propagation  and masking in networks.

\section*{Acknowledgments}

This work was supported in part by the European Union through European Social Fund in the frames of the ``Information and Communication Technologies (ICT) programme'' (``ITA-IoIT'' topic), by the Estonian Research Council grant PUT PRG1467 ``CRASHLES'' and by Estonian-French PARROT project ``EnTrustED''.

\bibliographystyle{IEEEtran}
\bibliography{ref, bib/nns, bib/axc, bib/tools, bib/mop}

\end{document}